%% file: arxiv.tex
\documentclass{article}
\usepackage{arxiv,times}
\usepackage{amsmath,amssymb,amsfonts}
\usepackage{graphicx,booktabs,multirow,array}
\usepackage[table]{xcolor}
\usepackage{pifont}
\usepackage{url}
\usepackage{hyperref}

\hypersetup{colorlinks=true,linkcolor=blue,citecolor=blue,urlcolor=blue,
  pdftitle={What Matters for Latent Actions in Robot Learning},
  pdfauthor={Xizhou Bu, Qingda Hu, Lei Zhou, Lingfeng Zhang, Yingbo Tang, Zihao Liu, Xinyi Tao, Zhiqiang Ma, Qingqiu Huang, Chufeng Tang, Hongbo Wang, Jing Zhang, Jiayi Ma, Hangjun Ye, Wei Li, Xiaoshuai Hao}}

\newcommand{\circB}{\textcolor{blue!70!black}{\ding{108}}}
\newcommand{\circC}{\textcolor{red!70!black}{\ding{108}}}
\newcommand{\best}[1]{\textbf{\textcolor{red}{#1}}}

\definecolor{designI}{RGB}{248,251,255}
\definecolor{designIHeader}{RGB}{235,244,255}
\definecolor{designII}{RGB}{255,252,246}
\definecolor{designIIHeader}{RGB}{255,241,222}
\input{sec/research_questions}
\title{What Matters for Latent Actions in\\ Robot Learning}
\author{%
Xizhou~Bu\thanks{Xizhou Bu, Qingda Hu, Hongbo Wang, Wei Li are with the College of Intelligent Robotics and Advanced Manufacturing, Fudan University, Shanghai, 200433, China (e-mail: \mbox{\href{mailto:xzbu24@m.fudan.edu.cn}{\nolinkurl{xzbu24@m.fudan.edu.cn}}}, \mbox{\href{mailto:qdhu24@m.fudan.edu.cn}{\nolinkurl{qdhu24@m.fudan.edu.cn}}}, \mbox{\href{mailto:wanghongbo@fudan.edu.cn}{\nolinkurl{wanghongbo@fudan.edu.cn}}}, \mbox{\href{mailto:fd_liwei@fudan.edu.cn}{\nolinkurl{fd_liwei@fudan.edu.cn}}})}\textsuperscript{,*},
Qingda~Hu\textsuperscript{1,*},
Lei~Zhou\thanks{Xiaoshuai Hao, Hangjun Ye, Lei Zhou and Yingbo Tang are with the Xiaomi EV, Beijing, 100085, China (e-mail: \mbox{\href{mailto:haoxiaoshuai714@163.com}{\nolinkurl{haoxiaoshuai714@163.com}}}, \mbox{\href{mailto:yehangjun@xiaomi.com}{\nolinkurl{yehangjun@xiaomi.com}}}, \mbox{\href{mailto:e0669578@u.nus.edu}{\nolinkurl{e0669578@u.nus.edu}}}, \mbox{\href{mailto:tangyingbo197@gmail.com}{\nolinkurl{tangyingbo197@gmail.com}}}).}\textsuperscript{,*},
Lingfeng~Zhang\thanks{Lingfeng Zhang is with the Tsinghua Shenzhen International Graduate School, Tsinghua University, Shenzhen, 518055, China (e-mail: \mbox{\href{mailto:lfzhang715@gmail.com}{\nolinkurl{lfzhang715@gmail.com}}}).},
Yingbo~Tang\textsuperscript{2},
Zihao~Liu\thanks{Zihao Liu, Xinyi Tao, Zhiqiang Ma are with the Research Center for Intelligent Robotics, School of Astronautics, Northwestern Polytechnical University, Xi'an, 710072, China (e-mail: \mbox{\href{mailto:2536890272carl@gmail.com}{\nolinkurl{2536890272carl@gmail.com}}}, \mbox{\href{mailto:taoxinyi1016@qq.com}{\nolinkurl{taoxinyi1016@qq.com}}}, \mbox{\href{mailto:mzq870130@163.com}{\nolinkurl{mzq870130@163.com}}})},
Xinyi~Tao\textsuperscript{4},
Zhiqiang~Ma\textsuperscript{4},
Qingqiu~Huang\thanks{Jing Zhang, Jiayi Ma, and Qingqiu Huang are with the School of Artificial Intelligence, Wuhan University, Wuhan, 430072, China (e-mail: \mbox{\href{mailto:jingzhang.cv@gmail.com}{\nolinkurl{jingzhang.cv@gmail.com}}}, \mbox{\href{mailto:jyma2010@gmail.com}{\nolinkurl{jyma2010@gmail.com}}}, \mbox{\href{mailto:draco.huang@morphi.com}{\nolinkurl{draco.huang@morphi.com}}}).},
Chufeng~Tang\thanks{Chufeng Tang is with Morphi Intelligence Technology Co., Ltd., Shenzhen, 518063, China (e-mail: \mbox{\href{mailto:felix.tang@morphi.com}{\nolinkurl{felix.tang@morphi.com}}}).},
Hongbo~Wang\textsuperscript{1},
Jing~Zhang\textsuperscript{5},
Jiayi~Ma\textsuperscript{5},
Hangjun~Ye\textsuperscript{2},
Wei~Li\textsuperscript{1,$\dagger$},
Xiaoshuai~Hao\textsuperscript{2,$\dagger$}\\[0.5em]
{\normalfont\small \textsuperscript{*}Equal contribution. \textsuperscript{$\dagger$}Corresponding authors.}}
\iclrfinalcopy
\begin{document}

\maketitle
\begin{abstract}
Latent Action Models (LAMs) have emerged as a promising paradigm for enabling robot learning to leverage large-scale unlabeled videos through latent actions that serve as compact surrogates for physical actions. Despite rapid progress, research on LAM remains highly fragmented, with existing methods evaluating different design choices in isolation under inconsistent experimental settings, making it difficult to identify the factors that truly determine downstream robotic manipulation performance. In this work, we present the first comprehensive empirical study of latent action learning for robotic manipulation. We unify representative LAM methods within a common autoencoding framework and systematically investigate 41 LAM design choices across three dimensions, including latent action modeling paradigms, learning objectives and regularization methods, and latent action integration strategies. We further examine four proxy metrics for evaluating latent action quality and assess their ability to reliably predict downstream robotic manipulation performance. Extensive experiments on three widely used benchmarks provide strong empirical evidence that fine-tuning vision-language model (VLM) backbones with latent actions provides a stronger initialization for downstream policy learning, with further validation on real-world robot manipulation tasks. Our project page is available at \href{https://carldegio.github.io/latent_action.github.io/}{LAM.github.io}.

\end{abstract}
\input{sec/introduction}
\input{sec/methodology}
\input{sec/experiments}
\input{sec/limitation}

\label{main-text-end}
\bibliography{sec/reference}
\bibliographystyle{iclr2027_conference}
\clearpage
\appendix
\input{sec/relatedwork}
\input{sec/methodology_appendix}
\input{sec/experiments_appendix}

\end{document}

%% file: sec/research_questions.tex
\newcommand{\questionone}{Which latent action modeling paradigms produce the highest-quality latent \mbox{action representations?}}
\newcommand{\questiontwo}{How do regularization methods and their regularization strengths affect latent action~learning?}
\newcommand{\questionthree}{How should latent actions be integrated into physical action~prediction?}
\newcommand{\questionfour}{Are proxy metrics for latent actions predictive of downstream robotic manipulation~performance?}
\newcommand{\questionfive}{What is the most suitable latent action dimensionality, and is latent action normalization~necessary?}
\newcommand{\questionsix}{Does fine-tuning the VLM backbone with latent actions exhibit scaling laws in robotic~manipulation?}

%% file: sec/introduction.tex
\section{Introduction}
Actions drive the transition dynamics of the physical world, yet action modeling fundamentally relies on large-scale robot action data \citep{kang2026x, pertsch2025fast, dong2026actioncodec}. Unfortunately, collecting such data remains prohibitively costly and difficult, limiting the development of scalable action representations. Although embodied foundation models have advanced rapidly in recent years, much of this progress has been inherited from advances in adjacent foundation model paradigms rather than from corresponding advances in action modeling. In particular, vision-language-action (VLA) models \citep{zitkovich2023rt,black2024pi_0,kim2024openvla} have inherited advances in vision-language models (VLMs) \citep{bai2025qwen3,beyer2024paligemma,karamcheti2024prismatic}, while world action models (WAMs) \citep{bi2025motus,ye2026world,agarwal2026cosmos} have similarly benefited from recent progress in video generation models (VGMs) \citep{wan2025wan,ali2025world,zheng2024open}. These foundation models are enabled by web-scale image-text and video corpora, whose scale and diversity exceed those of available robot action data by several orders of magnitude. Consequently, current embodied foundation models \citep{generalist2026gen1,agarwal2026cosmos} are typically initialized from large-scale pretrained VLMs or VGMs and subsequently adapted to robotic manipulation using comparatively limited robot action data \citep{ji2025robobrain,wu2025robocoin,o2024open}. Despite recent efforts have explored leveraging egocentric human videos \citep{li2025scalable,kareer2025emergence,yang2025egovla} and low-cost teleoperation systems such as universal manipulation interface (UMI) \citep{chi2024universal,liu2024fastumi,ohkawa2026yubi} to scale data collection, the scarcity of large-scale, high-quality robot action data remains unresolved. This imbalance between abundant perceptual data and scarce robot action data fundamentally limits the scalability of embodied foundation models and hinders the emergence of scaling behavior analogous to that observed in the training of \mbox{large language models \citep{kaplan2020scaling}}.

\begin{figure}[t!]
    \centering
    \includegraphics[width=0.99\linewidth]{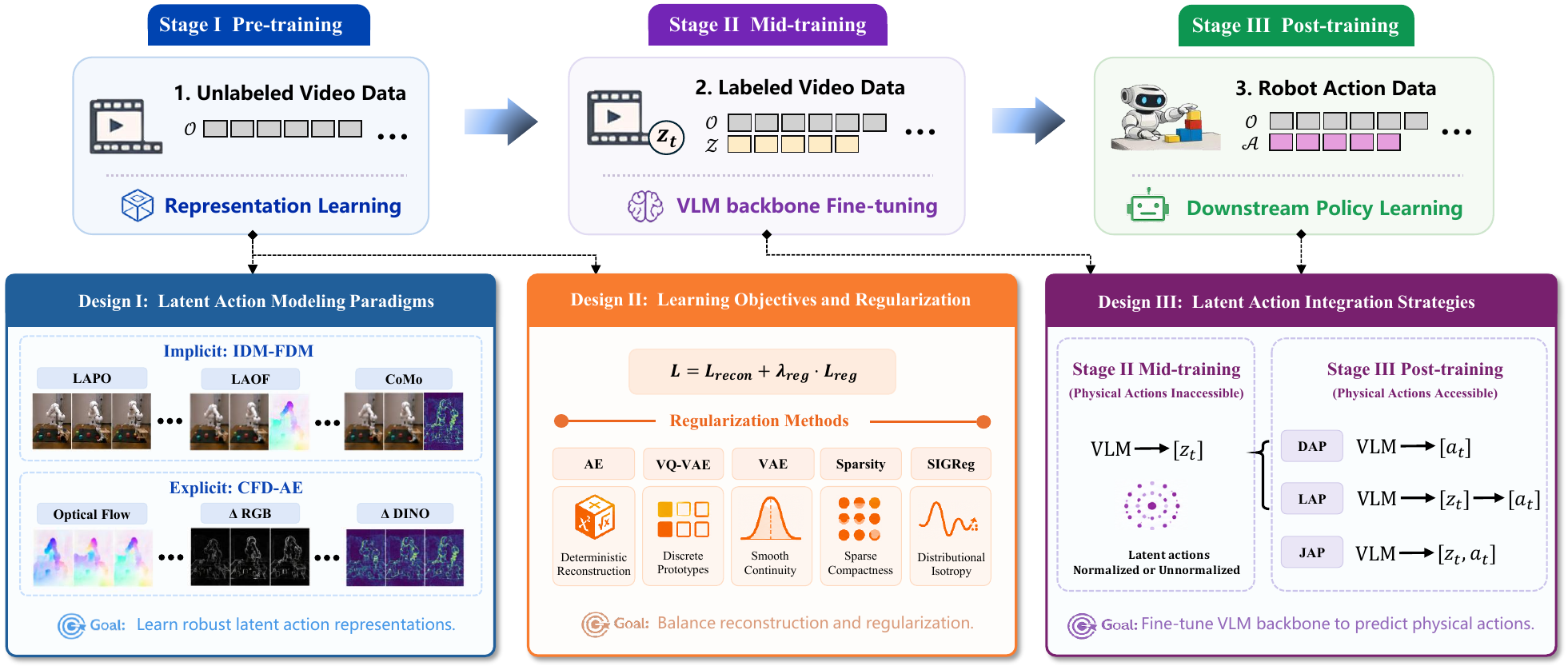}
    \caption{\looseness=-1\textbf{Overview of design choices across three dimensions for latent action learning.} Design I examines the impact of latent action modeling paradigms by comparing the implicit IDM-FDM framework with the explicit CFD-AE framework. Design II investigates how different regularization methods and their regularization strengths shape latent action representations and affect downstream policy learning. Design III evaluates the necessity of latent action normalization and investigates different strategies for integrating latent actions into physical action prediction. Stage I is trained on unlabeled video data consisting only of image frames, Stage II uses labeled video data paired with latent actions, and Stage III uses robot action data paired with physical actions. Further details of the three design studies and the three-stage training pipeline are provided in Section~\ref{sec:methodology} and Appendix~\ref{app:training_pipeline},~\mbox{respectively.}
    }
    \label{fig:main}
\end{figure}

\looseness=-1
A natural way to overcome this bottleneck is to exploit large-scale unlabeled videos to train latent action models (LAMs), which extract compact latent actions that serve as surrogates for physical actions \citep{edwards2019imitating,schmidt2024learning,liu2025stamo}. A representative framework is LAPO \citep{schmidt2024learning}, which learns latent action representations through a self-supervised autoencoding framework consisting of an inverse dynamics model and a forward dynamics model. Specifically, the inverse dynamics model encodes consecutive frames into a low-dimensional latent action that captures the transition dynamics, while the forward dynamics model predicts the future frame from the current frame conditioned on the inferred latent action. This paradigm enables scalable learning from abundant unlabeled video and has demonstrated promising performance in both VLA \citep{ye2025latent,bu2025univla,bu2025agibot} and WAM \citep{bi2025motus}. However, current research on LAM remains fragmented and lacks a coherent empirical understanding. Existing methods vary substantially in how latent action representations are learned from consecutive frames \citep{bu2026laof,yang2026learning,bi2025motus,liu2025stamo}, how they are regularized \citep{bu2025univla,gao2025adaworld,maes2026leworldmodel,garrido2026learning}, and how latent actions are integrated into physical action prediction \citep{gao2026dreamdojo,liu2025rdt,lin2026pixels}. Moreover, these design choices are typically proposed in isolation and evaluated under inconsistent experimental settings, making it unclear which factors fundamentally affect latent action learning and which merely have incidental effects. Another major obstacle is the lack of a reliable protocol for evaluating latent action quality. While downstream robotic manipulation performance provides the most faithful evaluation of latent action quality, obtaining such measurements requires completing the full three-stage training pipeline, as illustrated in Fig.~\ref{fig:main}, making iterative LAM development prohibitively expensive. As a result, prior work has relied on proxy metrics, mostly by training a lightweight multi-layer perceptron (MLP) probe during the pre-training stage to decode latent actions into physical actions and using the prediction error as a proxy for latent action quality \citep{schmidt2024learning,yang2026learning,bu2026laof,nie2026lary}. Yet, the validity of this proxy metric remains largely unverified, and whether it reliably predicts downstream robotic manipulation performance remains unclear.

In this work, we conduct a systematic empirical study of LAM design choices across three dimensions, including latent action modeling paradigms, learning objectives and regularization methods, and latent action integration strategies. Specifically, we investigate 41 design choices across the LIBERO, LIBERO-Plus, and RoboTwin2.0 benchmarks, with further validation on real-world robotic manipulation tasks using a Franka Panda robot. We further evaluate four proxy metrics for latent action quality, analyze their correlation with downstream robotic manipulation performance, and investigate the effect of expanding the Stage-II VLM fine-tuning corpus on downstream policy learning. Our contributions are summarized as follows:

(1) The original LAPO method remains a remarkably strong baseline when trained directly on raw data, while simple semantic feature differencing using off-the-shelf visual encoders is also sufficient to yield competitive latent action representations.

(2) We identify effective hyperparameter settings for different regularization methods, showing that properly tuned methods achieve comparable downstream performance. We further evaluate five latent action integration strategies and derive practical guidelines for effectively incorporating latent actions into physical action prediction.

(3) A latent action dimensionality of 32 achieves the highest average performance across the evaluated benchmarks. Additional latent action normalization generally does not improve performance when appropriate pre-training regularization is applied.

(4) FDM reconstruction metrics tend to provide more reliable indications of downstream robotic manipulation success rates than probe metrics. These proxy metrics are better suited to coarse-grained model selection than to fine-grained ranking and are insufficient for reliably identifying \mbox{the best-performing model.}

(5) Both simulation and real-world experiments demonstrate that fine-tuning VLM backbones with latent actions provides stronger initialization for downstream policy learning. Expanding the Stage-II VLM fine-tuning corpus with OXE data improves downstream robotic manipulation performance \mbox{across the evaluated benchmarks}.

%% file: sec/methodology.tex
\section{Methodology}
\label{sec:methodology}

\paragraph{Training Pipeline Setting}
\label{subsec:training_pipeline}
The overall framework consists of three stages: pre-training, mid-training, and post-training, corresponding to representation learning, VLM backbone fine-tuning, and downstream policy learning, respectively.
\textbf{Stage I (Pre-training).} This stage aims to learn a high-quality LAM from consecutive frames $(o_t,o_{t+1})$ to produce physically meaningful latent action representations that capture transition dynamics.
\textbf{Stage II (Mid-training).} The pretrained inverse dynamics model is employed as a latent action annotator to automatically generate latent actions for semantically conditioned video data, yielding triplets $(o_t,z_t,l)$. These latent actions are then used to fine-tune the VLM backbone, producing the embodied foundation model studied in this work.
\textbf{Stage III (Post-training).} The fine-tuned VLM backbone is subsequently optimized using a limited amount of robot action data $(o_t,a_t,l)$ for downstream policy learning.

\paragraph{Design I: Latent Action Modeling Paradigms}
We consider a set of representative methods for latent action modeling, which can be broadly divided into two paradigms, as illustrated in Fig.~\ref{fig:cfm}. The first is the inverse dynamics model--forward dynamics model (IDM-FDM) framework, which implicitly learns latent action representations through jointly trained inverse and forward dynamics models, including LAPO \citep{schmidt2024learning} and its variants, LAOF \citep{bu2026laof} and CoMo \citep{yang2026learning}. The second is the consecutive-frame differences autoencoding (CFD-AE) framework, which explicitly learns latent action representations by autoencoding consecutive-frame differences $\Delta_t = \Delta(o_{t+1}, o_t)$. In this work, $\Delta_t$ is instantiated as $\Delta$RGB, $\Delta$DINO, or \mbox{optical flow.}

\paragraph{Design II: Learning Objectives and Regularization}
A systematic investigation of latent action learning requires understanding how different learning objectives and regularization methods shape latent action representations, as existing methods lead to substantial differences in the structure, compactness, and controllability of the learned representations. To provide a unified analysis, we consider several representative learning objectives and regularization methods, including AE \citep{bank2023autoencoders}, VAE \citep{kingma2013auto}, VQ-VAE \citep{van2017neural}, Sparsity \citep{garrido2026learning}, and SIGReg \citep{balestriero2025lejepa}.

\paragraph{Design III: Latent Action Integration Strategies}
We consider three representative latent action integration strategies, namely direct action prediction (DAP), latent-to-action prediction (LAP), and joint latent-action prediction (JAP),  which differ in whether latent actions are used only to fine-tune the VLM backbone, explicitly serve as an intermediate control representation, or are jointly predicted with physical actions. To evaluate whether a jointly fine-tuned backbone can be used with different downstream action heads, we additionally design JAP-DAP and JAP-LAP. Specifically, in Stage III, the VLM backbone is first jointly fine-tuned with both latent actions and physical actions following the JAP setting. The trained action head is then discarded, and a new DAP or LAP \mbox{head is trained from scratch.}

%% file: sec/experiments.tex
\section{Experiments}

\label{sec:experiments}

\input{sec/results_summary}

Our experiments are organized into three dimensions, each comprising multiple design choices that examine different aspects of latent action learning. Design I focuses on latent action modeling paradigms with 7 choices. Design II covers learning objectives and regularization, with 21 choices spanning different regularization methods and their corresponding strengths. Design III studies latent action integration, with 15 choices covering latent action normalization, dimensionality, and integration strategies. After removing overlapping configurations, the three dimensions yield 41 unique design choices in total. Additional results and analyses are provided in Appendix~\ref{app:experiments}.

\subsection{Experimental Setup}

\label{sec:experimental_setting}

\noindent\textbf{Datasets.} In Stage I and Stage II, we use only raw video data without accessing robot action annotations. To investigate the effect of expanding the Stage-II VLM fine-tuning corpus, we construct a heterogeneous video-only corpus containing approximately 59M frames, whose composition is illustrated in Fig.~\ref{fig:scalelaw}. Unless otherwise specified, all experiments in Stages I and II use the full data mixture, while large-scale ablation studies of Design I–III are conducted on the \texttt{Robotwin} and \texttt{liberoplus} datasets.

\noindent\textbf{Implementation Details.} In Stage I, we implement LAM as a 700M spatiotemporal Transformer \citep{bruce2024genie}. In Stages II and III, we adopt OpenVLA-OFT \citep{kim2025fine} as the baseline following the implementation of StarVLA \citep{community2026starvla}, and use Qwen3-VL-4B \citep{bai2025qwen3} as the VLM backbone. To eliminate confounding factors when evaluating the effect of the VLM fine-tuned with latent actions on downstream robotic manipulation performance in Stage III, both training and evaluation use only front-view observations, without robot joint states, wrist-view observations, or any data augmentation. All experiments are conducted on 8 NVIDIA H200 GPUs.

\noindent\textbf{Benchmark.} Our simulation experiments cover two control settings: 7-DoF single-arm end-effector delta control and 14-DoF dual-arm joint position control. Unless otherwise specified, all ablation results are reported as the average success rate across 3 random seeds. Specifically, LIBERO results are averaged over its 4 task suites, LIBERO-Plus results are averaged over the same 4 task suites under the \texttt{RobotInit} perturbation, and RoboTwin2.0 results are averaged over 12 selected tasks, with each task evaluated under both easy and hard modes. Here, zero-shot evaluation on LIBERO-Plus refers to deployment under \texttt{RobotInit} after Stage-III policy training on the \texttt{Libero} datasets only; \texttt{liberoplus} videos are used in Stages I and II.

\subsection{\questionone~(\textbf{Q1})}

\label{sec:q2_cfm}

As shown in Table~\ref{tab:main_results}, we compare seven choices of latent action modeling paradigms across the LIBERO, LIBERO-Plus, and RoboTwin2.0 benchmarks, with each choice evaluated using five latent action integration strategies, including DAP, LAP, JAP, JAP-DAP, and JAP-LAP. Here, \textit{Avg.} denotes the average performance across the five latent action integration strategies. All methods are evaluated with latent action dimensionality $d_z=64$ and regularization strength $\lambda_{reg}=1\times10^{-6}$. Averaging over all benchmarks, the overall ranking is LAPO (0.7327) $>$ $\Delta$DINO (0.7280) $>$ CoMo (0.7167) $>$ $\Delta$RGB (0.6973) $>$ LAOF (0.6907) $>$ SEA-RAFT (0.6433) $>$ RAFT (0.6427). LAPO achieves the best overall performance, suggesting that it learns more robust latent action representations for robotic manipulation. Notably, LAPO requires no additional preprocessing and can be trained directly on raw data, highlighting its simplicity and effectiveness.

With optical flow computed directly from the raw data and without any additional filtering, SEA-RAFT and RAFT consistently achieve the lowest performance across all three benchmarks, even underperforming the simple $\Delta$RGB baseline. Moreover, LAOF extends LAPO by incorporating optical flow as an additional constraint, yet this modification consistently degrades performance.

Furthermore, $\Delta$DINO consistently outperforms $\Delta$RGB across all benchmarks. This result suggests that the representation space in which consecutive-frame differences are computed plays an important role in latent action modeling. Notably, $\Delta$DINO achieves overall performance close to that of LAPO and even outperforms LAPO on LIBERO, suggesting that a strong pretrained semantic prior can itself provide an effective inductive bias for latent action learning in robotic manipulation. Finally, LAPO consistently outperforms CoMo across all three benchmarks. The primary difference between the two methods is that CoMo replaces the future-frame input to the IDM with semantic feature differences to mitigate potential information leakage, but this modification consistently degrades downstream robotic \mbox{manipulation performance.}

\begin{figure}[t]
	\centering
    \includegraphics[width=\linewidth]{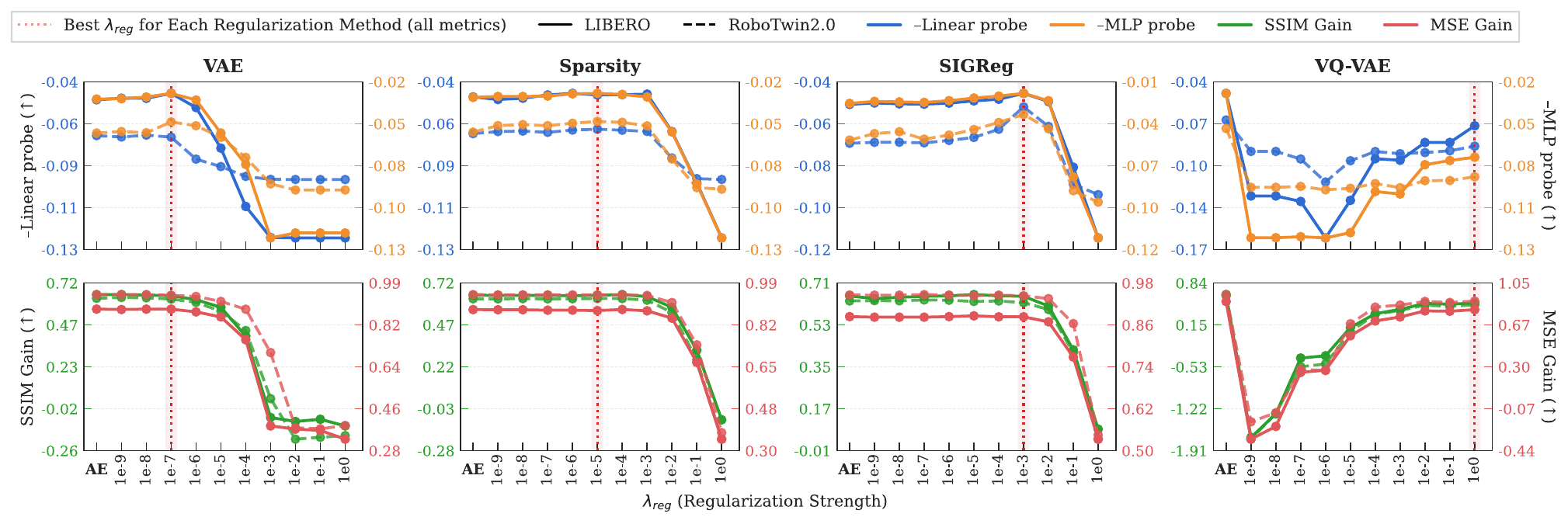}
    \caption{\textbf{Effect of regularization strength $\lambda_{reg}$ on latent action learning.} Columns show four regularization methods, with AE as the unregularized baseline. The top and bottom rows show probe and reconstruction metrics (Section~\ref{sec:q1_proxy_metrics}), respectively, on color-matched vertical axes. Solid and dashed curves denote LIBERO and RoboTwin2.0; dotted vertical lines mark selected strengths.}
\label{fig:lambda_selection}
\end{figure}

\subsection{\questiontwo~(\textbf{Q2})}

\label{sec:q3_regularization}

We find that the choice of regularization type has a relatively limited impact on downstream robotic manipulation performance, whereas the regularization strength plays a much more critical role in determining latent action quality, as shown in Fig.~\ref{fig:lambda_selection}. Unless otherwise specified, all Design-II results are obtained using LAPO with a latent action dimensionality of $d_z=64$.

For continuous latent actions, VAE, Sparsity, and SIGReg achieve comparable downstream performance on both LIBERO and RoboTwin2.0, as shown by the Design-II results in Table~\ref{tab:main_results}. For example, on LIBERO, their average success rates range only from 0.908 to 0.915. VQ-VAE achieves the strongest zero-shot generalization on LIBERO-Plus, with an average success rate of 0.517, outperforming AE (0.445), VAE (0.433), Sparsity (0.468), and SIGReg (0.467). This observation suggests that discretizing latent actions may improve robustness under \mbox{distribution shifts.}

Weak-to-moderate regularization is generally the safest choice for continuous latent actions, and we recommend $\lambda_{\mathrm{reg}}=10^{-7}$ for VAE, $10^{-5}$ for Sparsity, and $10^{-3}$ for SIGReg. For discrete latent actions, $\lambda_{\mathrm{reg}}=1$ is recommended for VQ-VAE. VAE is thus the recommended default when simplicity and effectiveness are prioritized, while VQ-VAE is preferred when zero-shot generalization is the primary concern.

\begin{figure}[t]
	\centering
    \includegraphics[width=0.98\linewidth]{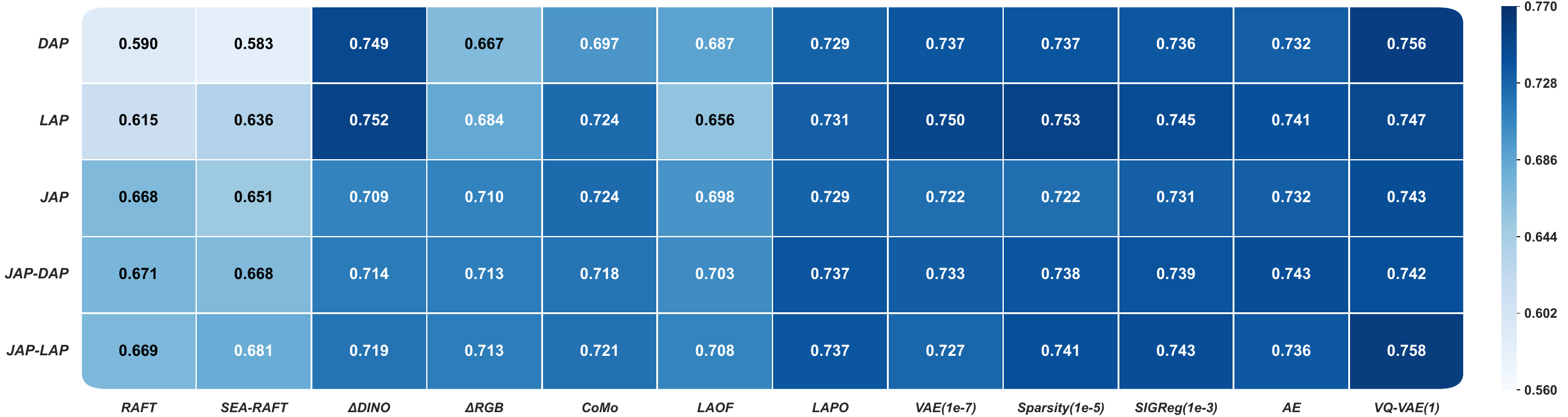}
    \caption{\textbf{Performance comparison of different latent action integration strategies.} The horizontal axis shows seven latent action modeling paradigms and five regularization methods. The vertical axis shows five latent action integration strategies. Each cell reports the mean success rate across LIBERO, LIBERO-Plus, and RoboTwin2.0.}
\label{fig:actionhead}
\end{figure}

\subsection{\questionthree~(\textbf{Q3})}

\label{sec:q4_action_head}

We consider two settings for incorporating latent actions into downstream policy learning: Physical Action Inaccessible and Physical Action Accessible. The distinction is whether physical action annotations are available during VLM backbone fine-tuning with latent actions. DAP and LAP are evaluated in the former setting, and JAP, JAP-DAP, and JAP-LAP in the latter. Both settings use robot action annotations for downstream policy learning.

\noindent\textbf{Physical Action Inaccessible.} In this setting, we compare DAP and LAP, both of which use only latent actions to fine-tune the VLM backbone in Stage II. Unlike DAP, LAP preserves latent actions as an intermediate representation between VLM features and physical actions. As shown in Fig.~\ref{fig:actionhead}, LAP outperforms DAP in 10 of the 12 configurations after averaging over the three benchmarks for each configuration. Equally weighting the 12 configurations and three benchmarks in Table~\ref{tab:main_results}, LAP achieves a higher overall mean success rate than DAP (0.7114 vs. 0.7000). In addition, its mean is close to that of JAP (0.7115), which uses physical action annotations during joint fine-tuning. Therefore, we recommend LAP when only video data are available for Stage II.

\noindent\textbf{Physical Action Accessible.} In this setting, we compare JAP and its two variants, JAP-DAP and JAP-LAP, all of which jointly use latent actions and physical actions to fine-tune the VLM backbone. JAP jointly predicts latent and physical actions from shared VLM features. To evaluate whether a jointly fine-tuned backbone can be used with different downstream action heads, we introduce JAP-DAP and JAP-LAP. Both variants retain the jointly fine-tuned backbone, discard the joint prediction head, and train a new DAP or LAP head from scratch. Across the same 12 configurations and three benchmarks, JAP-LAP achieves the highest overall mean success rate among these three strategies (0.7212), followed by JAP-DAP (0.7181) and JAP (0.7115). When robot action annotations are available for joint fine-tuning, JAP-LAP is the recommended choice based on its overall mean performance; the best strategy for an individual configuration or benchmark can differ.

\begin{figure}[t]
    \centering
    \includegraphics[width=\linewidth]{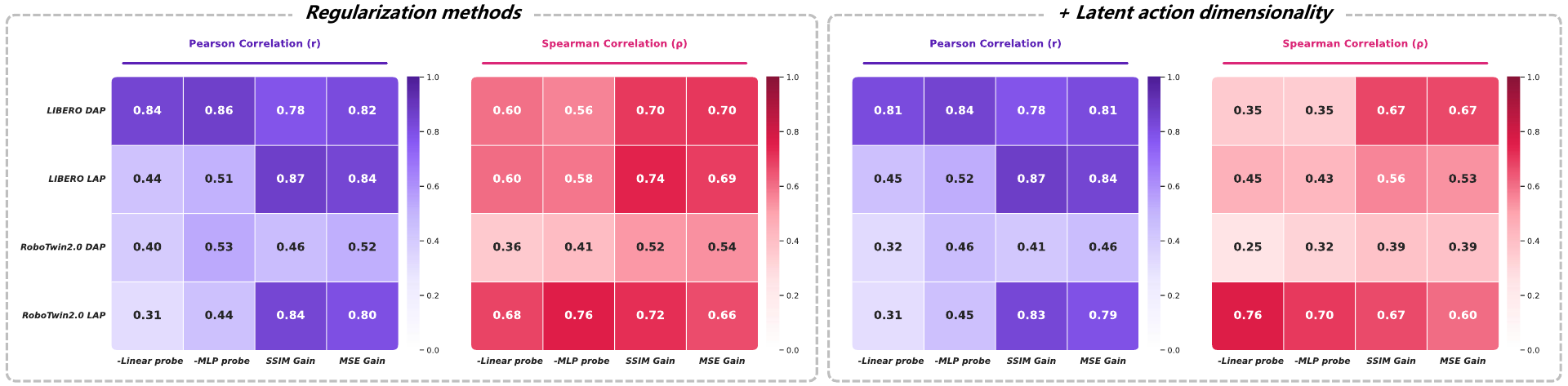}
    \caption{\textbf{Correlation between proxy metrics and downstream robotic manipulation performance.} Within each panel, columns correspond to the four proxy metrics and rows to benchmark--strategy combinations (LIBERO/RoboTwin2.0 with DAP/LAP). The purple and red panels report the Pearson correlation ($r$) and Spearman correlation ($\rho$), respectively.}
    \label{fig:latentaction_metrics}
\end{figure}

\subsection{\questionfour~(\textbf{Q4})}

\label{sec:q1_proxy_metrics}

\noindent\textbf{Proxy Metrics.} We evaluate latent action quality using four proxy metrics and analyze their correlation with downstream robotic \mbox{manipulation performance.} $\mathrm{Linear\ probe}$ and $\mathrm{MLP\ probe}$ evaluate the information about physical actions preserved in latent actions by measuring the accuracy of physical action recovery using lightweight action decoders. Specifically, the $\mathrm{Linear\ probe}$ employs a linear action decoder with architecture $[d_z,d_a]$, whereas the $\mathrm{MLP\ probe}$ adopts a nonlinear action decoder with architecture $[d_z,128,128,d_a]$. To ensure a fair comparison across different latent action representations, both probes are trained under the same settings with a learning rate of $1\times10^{-3}$, 50 training steps, and a global batch size of 256.

Lower $\mathcal{L}_{\mathrm{act}}$, the physical action prediction loss, indicates that latent actions preserve more information relevant to physical action prediction. Accordingly, we report $-\mathrm{Linear\ probe}$ and $-\mathrm{MLP\ probe}$ in the correlation analysis so that larger values consistently indicate higher latent action quality across all proxy metrics. $\mathrm{SSIM\ Gain}$ and $\mathrm{MSE\ Gain}$ are normalized reconstruction metrics that quantify the improvement of future-frame reconstruction over directly using the current frame as the prediction. Specifically, $\mathrm{SSIM\ Gain}$ is defined as $[\mathrm{SSIM}(o_{t+1},\hat{o}_{t+1})-\mathrm{SSIM}(o_t,o_{t+1})]/[1-\mathrm{SSIM}(o_t,o_{t+1})]$, where $\mathrm{SSIM}$ denotes the structural similarity index and $\hat{o}_{t+1}$ denotes the FDM-predicted future frame. $\mathrm{MSE\ Gain}$ is defined as $[\mathrm{MSE}(o_t,o_{t+1})-\mathrm{MSE}(o_{t+1},\hat{o}_{t+1})]/\mathrm{MSE}(o_t,o_{t+1})$. Larger $\mathrm{SSIM\ Gain}$ and $\mathrm{MSE\ Gain}$ indicate that the learned latent actions provide more informative transition dynamics, enabling the FDM to transform the current frame into the future frame rather than simply copying the current frame.

Our correlation analysis suggests that the proposed proxy metrics provide useful signals of downstream robotic manipulation performance, but they should not be regarded as a complete replacement for task evaluation. As shown in Fig.~\ref{fig:latentaction_metrics}, the Pearson and Spearman correlations are positive but vary across the evaluated settings, and proxy scores do not fully preserve the downstream ordering of configurations, indicating that the proxy metrics are more reliable for coarse-grained model screening than for fine-grained model ranking. In other words, the proxy metrics can help identify promising configurations, but downstream task evaluation remains necessary for model selection.

Overall, FDM reconstruction metrics ($\mathrm{SSIM\ Gain}$ and $\mathrm{MSE\ Gain}$) tend to reflect downstream robotic manipulation success rates more consistently than probe metrics ($\mathrm{-Linear\ probe}$ and $\mathrm{-MLP\ probe}$). Reconstruction metrics assess future-frame prediction, whereas probe metrics assess physical action recovery through an additional decoder. We also find that the correlations of all proxy metrics deteriorate when comparing models with different latent action dimensionalities. We therefore recommend using these proxy metrics to compare models with the same latent action dimensionality rather than across different dimensionalities. As shown in Fig.~\ref{fig:training_process}, it can be observed that LIBERO under end-effector delta control consistently yields lower probe losses than RoboTwin2.0 under joint position control over the training steps shown.

\begin{figure}[t]
    \centering
    \includegraphics[width=\linewidth]{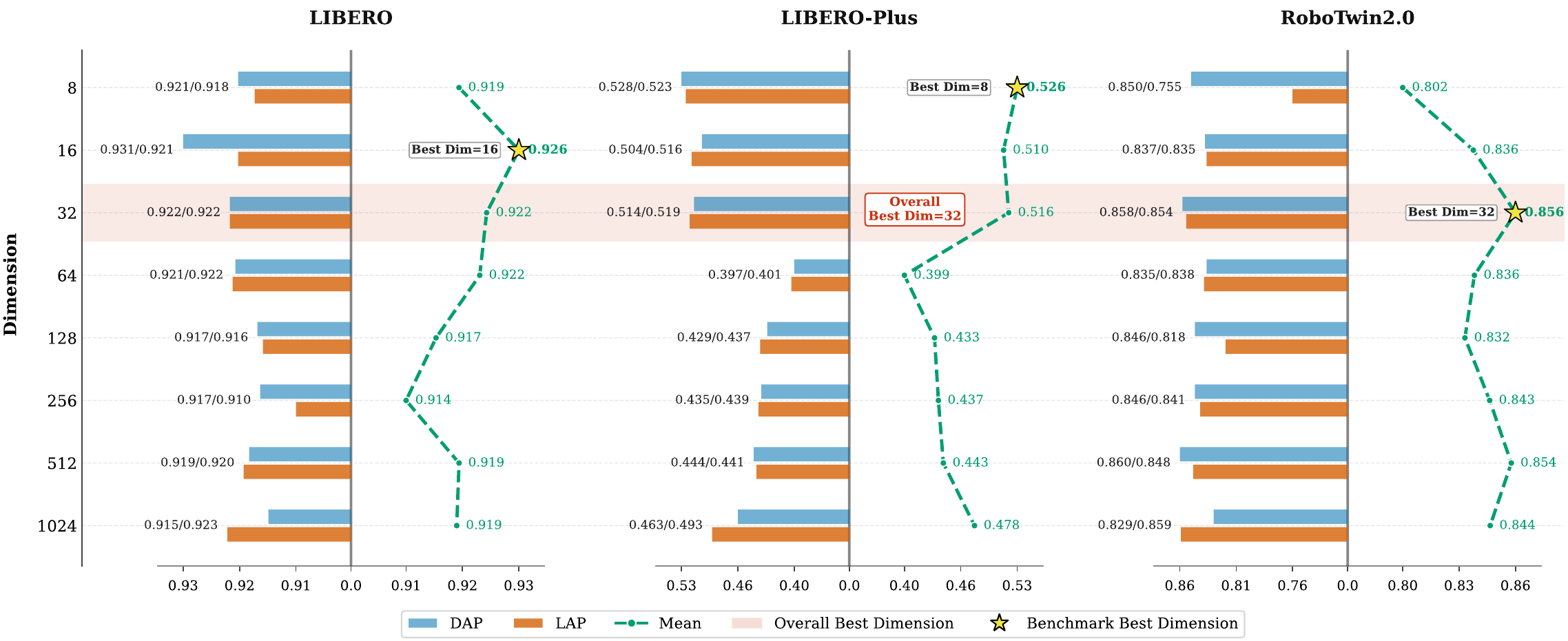}
    \caption{\textbf{Effect of latent action dimension on downstream robotic manipulation performance.} Blue (DAP) and orange (LAP) bars show success rates across latent action dimensions; the green dashed curve shows their mean. Stars mark the best dimension per benchmark; shading marks the best dimension averaged across benchmarks. Results use LAPO with VAE regularization ($10^{-6}$).}
    \label{fig:latentdims}
\end{figure}

\subsection{\questionfive~(\textbf{Q5})}

\label{sec:q5_latent_action_dimension}

For both 7-DoF single-arm and 14-DoF bimanual robot control, a latent action dimensionality of $d_z=32$ provides the best overall trade-off, as shown in Fig.~\ref{fig:latentdims}. On LIBERO, a 7-DoF single-arm benchmark, the downstream robotic manipulation performance is largely saturated with respect to the latent dimensionality. Although the highest average success rate is achieved with $d_z=16$ (0.926), the performance of $d_z=32$ is nearly identical (0.922). A similar trend is observed on LIBERO-Plus, which also evaluates 7-DoF single-arm manipulation while introducing diverse environmental perturbations to better assess policy robustness. More compact latent representations perform slightly better, with $d_z=8$ achieving the highest average success rate of 0.526, whereas $d_z=32$ remains highly competitive at 0.516. In contrast, RoboTwin2.0 evaluates 14-DoF bimanual manipulation, which requires coordinated dual-arm motions and therefore places higher demands on representation capacity. Under this setting, $d_z=32$ achieves the best overall performance (0.856), substantially outperforming $d_z=8$ (0.802).

We further investigate the effect of normalizing latent actions for VLM backbone fine-tuning in Stage II. As shown in Fig.~\ref{fig:denorm}, when the latent space is appropriately regularized, using unnormalized latent actions generally yields better downstream robotic \mbox{manipulation performance.} Across 33 method and benchmark combinations, 28 exhibit performance improvements without normalization, with an average gain of $+0.0115$. The largest improvement is observed on LIBERO-Plus ($+0.0160$), indicating that preserving the original scale of latent actions is particularly beneficial under challenging \mbox{distribution shifts.}

\begin{figure}[t]
    \centering
    \includegraphics[width=\linewidth]{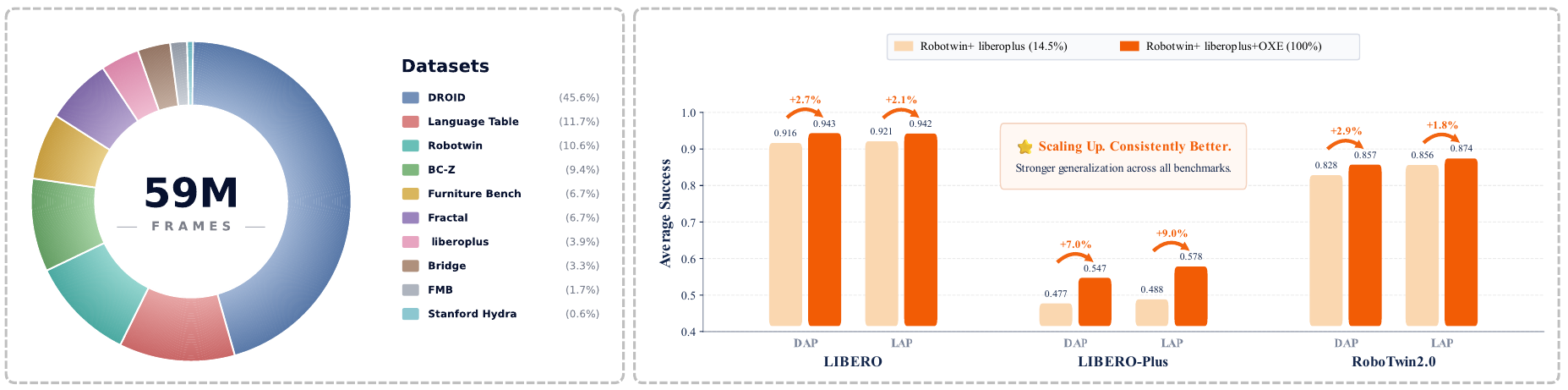}
    \caption{\textbf{Effect of Stage-II corpus expansion on downstream robotic manipulation performance.} The left panel shows the full corpus composition; \texttt{OXE} includes all datasets except \texttt{Robotwin} and \texttt{liberoplus}. The right panel reports average DAP and LAP success rates by benchmark. Light and dark orange bars denote Stage-II training without and with \texttt{OXE}, respectively.}
    \label{fig:scalelaw}
\end{figure}

\subsection{\questionsix~(\textbf{Q6})}

\label{sec:q6_scalelaw}

Expanding the Stage-II fine-tuning corpus with OXE data improves downstream robotic manipulation performance while keeping Stage III unchanged. As shown in Fig.~\ref{fig:scalelaw}, expanding the Stage-II fine-tuning corpus with OXE data improves performance across the evaluated benchmarks. The largest improvement is observed on LIBERO-Plus, with gains of up to 9.0 percentage points, suggesting that expanding the fine-tuning corpus is particularly beneficial for this benchmark. In comparison, the improvements on LIBERO and RoboTwin2.0 are smaller but remain consistently positive, indicating that the benefit of this corpus expansion extends across the evaluated benchmarks. Moreover, both DAP and LAP consistently benefit from the expanded Stage-II fine-tuning corpus across the evaluated benchmarks.

\begin{figure}[t]
    \centering
    \includegraphics[width=\linewidth]{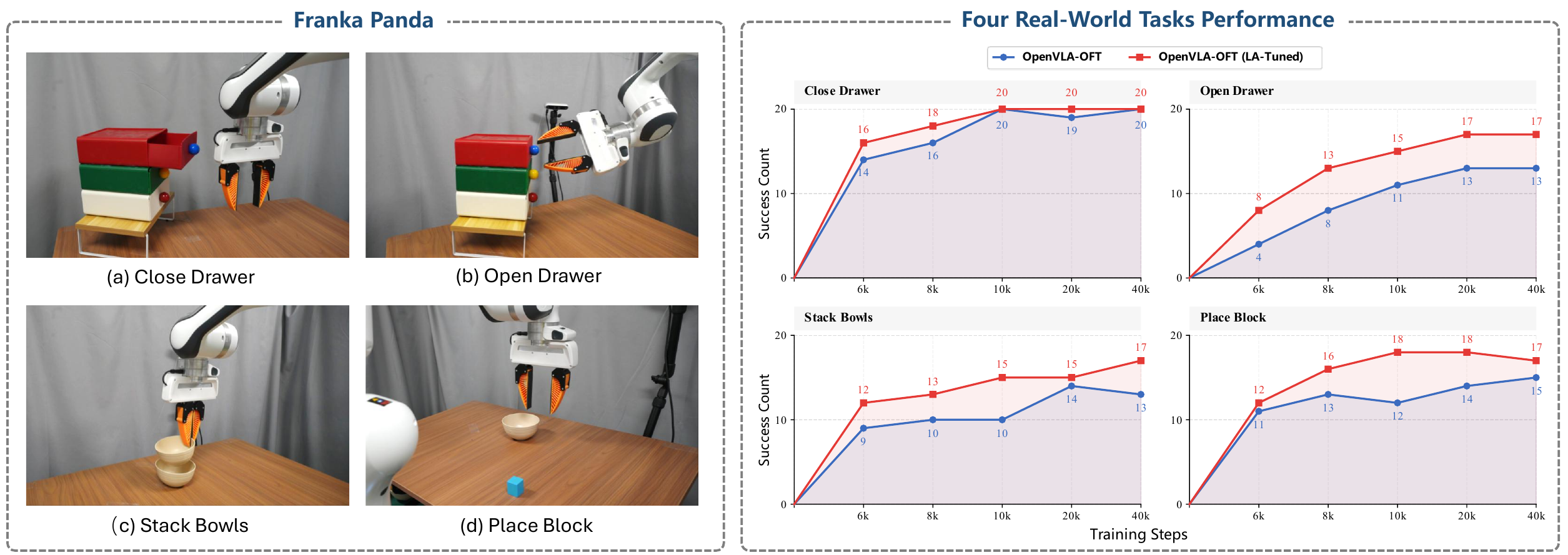}
    \caption{\textbf{Real-World Experiments.} The left panel shows the four tasks on the Franka Panda platform. The right panel plots success counts out of 20 trials against Stage-III training steps. Blue circles and red squares denote OpenVLA-OFT and OpenVLA-OFT (LA-Tuned), respectively.}
    \label{fig:real_world_tasks}
\end{figure}

\paragraph{Real-World Experiments.}

To further evaluate the effectiveness of fine-tuning the VLM backbone using latent actions in real-world environments, we conduct experiments on a 7-DoF single-arm Franka Panda platform equipped with a 1-DoF UMI gripper, as illustrated in the left panel of Fig.~\ref{fig:real_world_tasks}. Further experimental details are provided in Appendix~\ref{app:real_world_experiments}.

\looseness=-1
As shown in the right panel of Fig.~\ref{fig:real_world_tasks}, fine-tuning the VLM backbone with latent actions consistently improves real-world manipulation performance on the Franka Panda platform. Across all four tasks and five Stage III checkpoints, OpenVLA-OFT (LA-Tuned) achieves 317 successful rollouts out of 400, compared with 259/400 for the original OpenVLA-OFT baseline, improving the overall success rate from 64.75\% to 79.25\% (+14.5 percentage points, or +22.4\% relative improvement). The gains are observed on all task categories, including stacking, object placement, and articulated-object manipulation. Notably, the LA-Tuned backbone reaches a higher success rate earlier in Stage III training, achieving an average success rate of 85.0\% after only 10k Stage III training steps, already exceeding the baseline performance of 76.25\% obtained after 40k steps. These results indicate that fine-tuning the VLM backbone with latent actions provides a stronger initialization for understanding physical-world interactions, leading to faster Stage III learning and stronger real-world robotic \mbox{manipulation performance.}

%% file: sec/results_summary.tex
\begin{table}[t]
\centering
\caption{Performance comparison across LIBERO, LIBERO-Plus, and RoboTwin2.0. Entries report success rates. Avg. averages over the five integration strategies; red boldface marks the highest Avg. within each design group and benchmark.}
\label{tab:main_results}

\begingroup
\setlength{\tabcolsep}{1.5pt}
\renewcommand{\arraystretch}{1.5}
\newlength{\resultcolumnwidth}
\settowidth{\resultcolumnwidth}{JAP-DAP}
\newcolumntype{N}{w{c}{\resultcolumnwidth}}

\resizebox{\textwidth}{!}{
\begin{tabular}{l*{5}{N!{\kern\arrayrulewidth}}N|*{5}{N!{\kern\arrayrulewidth}}N|*{5}{N!{\kern\arrayrulewidth}}N}
\toprule

\multirow{2}{*}{Method}
& \multicolumn{6}{c|}{LIBERO}
& \multicolumn{6}{c|}{LIBERO-Plus}
& \multicolumn{6}{c}{RoboTwin2.0} \\
\cmidrule(lr){2-7} \cmidrule(lr){8-13} \cmidrule(lr){14-19}
& DAP & LAP & JAP & JAP-DAP & JAP-LAP & Avg.
& DAP & LAP & JAP & JAP-DAP & JAP-LAP & Avg.
& DAP & LAP & JAP & JAP-DAP & JAP-LAP & Avg. \\
\midrule

\rowcolor{designIHeader}
\multicolumn{19}{c}{\textit{Design I: Latent Action Modeling Paradigms}} \\

\rowcolor{designI}
RAFT
& 0.676 & 0.714 & 0.838 & 0.836 & 0.837 & 0.780
& 0.310 & 0.359 & 0.340 & 0.346 & 0.356 & 0.342
& 0.785 & 0.773 & 0.825 & 0.831 & 0.815 & 0.806 \\

\rowcolor{designI}
SEA-RAFT
& 0.673 & 0.764 & 0.847 & 0.868 & 0.858 & 0.802
& 0.289 & 0.339 & 0.336 & 0.315 & 0.351 & 0.326
& 0.787 & 0.804 & 0.769 & 0.820 & 0.833 & 0.802 \\

\rowcolor{designI}
$\Delta$RGB
& 0.853 & 0.860 & 0.899 & 0.894 & 0.896 & 0.880
& 0.352 & 0.382 & 0.397 & 0.388 & 0.413 & 0.387
& 0.795 & 0.809 & 0.834 & 0.856 & 0.830 & 0.825 \\

\rowcolor{designI}
$\Delta$DINO
& 0.933 & 0.923 & 0.913 & 0.920 & 0.918 & \best{0.921}
& 0.458 & 0.488 & 0.396 & 0.408 & 0.402 & 0.430
& 0.855 & 0.846 & 0.817 & 0.813 & 0.837 & 0.833 \\

\rowcolor{designI}
CoMo
& 0.907 & 0.912 & 0.910 & 0.903 & 0.908 & 0.908
& 0.372 & 0.399 & 0.405 & 0.414 & 0.396 & 0.397
& 0.811 & 0.862 & 0.858 & 0.838 & 0.858 & 0.845 \\

\rowcolor{designI}
LAOF
& 0.889 & 0.874 & 0.895 & 0.889 & 0.892 & 0.888
& 0.325 & 0.285 & 0.387 & 0.377 & 0.395 & 0.354
& 0.848 & 0.810 & 0.812 & 0.843 & 0.836 & 0.830 \\

\rowcolor{designI}
LAPO
& 0.928 & 0.919 & 0.907 & 0.898 & 0.900 & 0.910
& 0.397 & 0.430 & 0.449 & 0.452 & 0.455 & \best{0.437}
& 0.862 & 0.845 & 0.830 & 0.862 & 0.856 & \best{0.851} \\

\midrule

\rowcolor{designIIHeader}
\multicolumn{19}{c}{\textit{Design II: Learning Objectives and Regularization}} \\

\rowcolor{designII}
VAE(1e-7)
& 0.921 & 0.913 & 0.905 & 0.904 & 0.906 & 0.910
& 0.452 & 0.463 & 0.400 & 0.424 & 0.425 & 0.433
& 0.838 & 0.875 & 0.860 & 0.870 & 0.852 & \best{0.859} \\

\rowcolor{designII}
Sparsity(1e-5)
& 0.920 & 0.915 & 0.913 & 0.915 & 0.916 & \best{0.915}
& 0.471 & 0.501 & 0.459 & 0.451 & 0.461 & 0.468
& 0.822 & 0.844 & 0.795 & 0.848 & 0.846 & 0.831 \\

\rowcolor{designII}
SIGReg(1e-3)
& 0.910 & 0.909 & 0.902 & 0.909 & 0.913 & 0.908
& 0.481 & 0.478 & 0.462 & 0.442 & 0.469 & 0.467
& 0.817 & 0.847 & 0.829 & 0.865 & 0.849 & 0.841 \\

\rowcolor{designII}
AE
& 0.928 & 0.925 & 0.901 & 0.906 & 0.904 & 0.913
& 0.440 & 0.455 & 0.436 & 0.449 & 0.447 & 0.445
& 0.828 & 0.844 & 0.859 & 0.873 & 0.856 & 0.852 \\

\rowcolor{designII}
VQ-VAE(1)
& 0.920 & 0.910 & 0.905 & 0.907 & 0.909 & 0.910
& 0.537 & 0.534 & 0.494 & 0.498 & 0.522 & \best{0.517}
& 0.811 & 0.799 & 0.830 & 0.821 & 0.845 & 0.821 \\

\bottomrule
\end{tabular}
}

\endgroup
\end{table}

%% file: sec/limitation.tex
\section{Conclusion and Future Work}
This work presents a systematic empirical study of latent action modeling paradigms, regularization methods, and integration strategies. LAPO remains a strong baseline, and LAP is a practical choice when only video data are available for Stage II. Proxy metrics are better suited to coarse-grained model screening than to fine-grained ranking. Real-world experiments further demonstrate the benefits of latent-action tuning for downstream policy learning.

Future work will explore incorporating latent actions into the pre-training of vision-language or multimodal foundation models. Jointly learning semantic and action representations from scratch may enable stronger alignment between language semantics and physical actions. We also plan to leverage large-scale in-the-wild video data from online platforms, such as YouTube, to expand the diversity and scale of training data. Finally, we will investigate whether our findings generalize across a broader range of robotic platforms, including dexterous hands, \mbox{quadruped robots, and humanoid robots.}

%% file: sec/relatedwork.tex
\section{Related Work}
\label{app:relatedwork}

\subsection{Latent Action Learning from Consecutive Frames}
Different methods exploit different sources of motion signals, ranging from semantic feature differences \citep{liu2025stamo} and optical flow \citep{bi2025motus} to implicitly learned transition dynamics \citep{schmidt2024learning,bu2026laof,yang2026learning}. However, it remains unclear which method is most effective for learning high-quality latent action representations that support downstream policy learning. Existing methods can be broadly categorized into two paradigms. The first paradigm models implicit transition dynamics through jointly learned inverse and forward dynamics models (IDM-FDM) under a self-supervised autoencoding framework \citep{schmidt2024learning}. Building on this, LAOF \citep{bu2026laof} introduces optical flow as an additional supervision signal during latent action learning to improve robustness to environmental distractors, while CoMo \citep{yang2026learning} mitigates shortcut learning, where the IDM directly encodes future-frame information, by replacing the future-frame input with the difference between consecutive frames. Several subsequent methods \citep{chen2025villa,liang2025clam,li2025latbot,wang2026imitation} further incorporate physical actions or robot states to inject robot control priors into latent action representations, leading to improved downstream robotic manipulation performance. Nevertheless, existing robot datasets remain highly heterogeneous, as different robot platforms employ different control interfaces, action space definitions, and control dimensionalities. To avoid these embodiment-specific confounding factors and focus on the intrinsic properties of latent action learning, we adopt a video-only learning setting throughout this stage. The second paradigm, which we refer to as consecutive-frame differences autoencoding (CFD-AE), explicitly learns latent action representations using predefined motion signals. For example, StaMo \citep{liu2025stamo} computes semantic feature differences using an off-the-shelf visual encoder such as DINOv2 \citep{oquab2023dinov2}, whereas Motus \citep{bi2025motus} directly extracts dense optical flow. To enable a fair comparison with implicit IDM-FDM methods, we uniformly transform both semantic feature differences and optical flow into latent actions through a unified autoencoding pipeline, thereby aligning the latent action dimensionality while eliminating architectural differences across methods.

\subsection{Regularization in Latent Action Learning}
In the standard IDM-FDM framework, the IDM infers a latent action from a pair of consecutive frames, while the FDM predicts the future frame conditioned on the current frame and the inferred latent action. This framework introduces a causal leakage issue, as the IDM has direct access to the target future frame during self-supervised training, potentially leading to shortcut learning. As a result, the inferred latent actions may encode excessive future frame information rather than the underlying transition dynamics, or even degenerate into a direct feature of the future frame itself \citep{yang2026learning}. A common strategy to mitigate this issue is to introduce an information bottleneck into the latent action representations. Several methods adopt discretization through VQ-VAE \citep{van2017neural}, which has been widely used in latent action learning \citep{bruce2024genie,bu2025univla,ye2025latent,bu2025agibot,bjorck2025gr00t,chen2024igor,chen2025villa,chen2024moto,cui2024dynamo,collins2025amplify,zhang2026clap,lin2026pixels,lyu2026lafp}. By learning a finite set of discrete prototypes, VQ-VAE effectively constrains the information capacity of latent action representations. This design was originally motivated by relatively simple discrete-control domains, such as pixel-based games \citep{bruce2024genie}, where physical actions can be naturally represented as discrete tokens. For robotic manipulation involving complex continuous control, recent works have increasingly adopted continuous latent action representations based on VAE \citep{kingma2013auto}, which provide greater flexibility and smoother transitions \citep{nikulin2025latent,gao2025adaworld,gao2026dreamdojo,yang2026learning,bu2026laof,liang2025clam,li2025latbot,bi2025motus}. In this setting, latent action representations are regularized by matching their distribution to a Gaussian distribution through a Kullback-Leibler (KL) divergence \citep{kullback1951information} objective. 
Other regularization objectives have also been proposed to improve the statistical properties of latent action representations and mitigate representation collapse. For example, Sparsity \citep{garrido2026learning} introduces a more sophisticated regularization objective that encourages sparse and compact representations. In contrast, LeWorldModel \citep{maes2026leworldmodel} employs the Sketched-Isotropic-Gaussian Regularizer (SIGReg) \citep{balestriero2025lejepa}, a simpler approach that promotes distributional isotropy by encouraging latent action representations to follow an isotropic Gaussian distribution.

%% file: sec/methodology_appendix.tex
\section{Detailed Methodology}
\label{app:methodology}

This appendix supplements Section~\ref{sec:methodology} with notation, learning objectives, and training details.

\subsection{Problem Setting}
We consider a sequential decision-making setting in which, at each time step $t$, an agent receives a current frame $o_t\in\mathcal{O}$, executes a physical action $a_t\in\mathcal{A}$, and transitions to the future frame $o_{t+1}$, where $\mathcal{O}$ denotes the observation space and $\mathcal{A}\subseteq\mathbb{R}^{d_a}$ denotes the physical action space of dimension $d_a$. Latent action learning aims to learn a latent action representation $z_t\in\mathcal{Z}$ from consecutive frames that captures the underlying transition dynamics, where $\mathcal{Z}\subseteq\mathbb{R}^{d_z}$ denotes the latent action space of dimension $d_z$. To provide a unified formulation, we denote the consecutive-frame differences by $\Delta_t=\Delta(o_{t+1},o_t)$, where $\Delta(\cdot,\cdot)$ denotes a function that extracts differences from consecutive frames. Under this formulation, a trajectory is represented as $\tau={(o_t,a_t,z_t,o_{t+1},l)}_{t=1}^{T}$, where $l$ denotes the language instruction associated with the trajectory.

\subsection{Training Pipeline Setting}
\label{app:training_pipeline}
The three-stage pipeline follows Section~\ref{subsec:training_pipeline}. Stage I requires neither robot action data nor task-specific policy demonstrations, and can be trained entirely on large-scale video data, including videos collected through random exploration or from the web. The stage-specific objectives and action-head initialization protocols are detailed below.

\subsection{Latent Action Modeling Paradigms}
\begin{figure}[t!]
    \centering
    \includegraphics[width=0.99\linewidth]{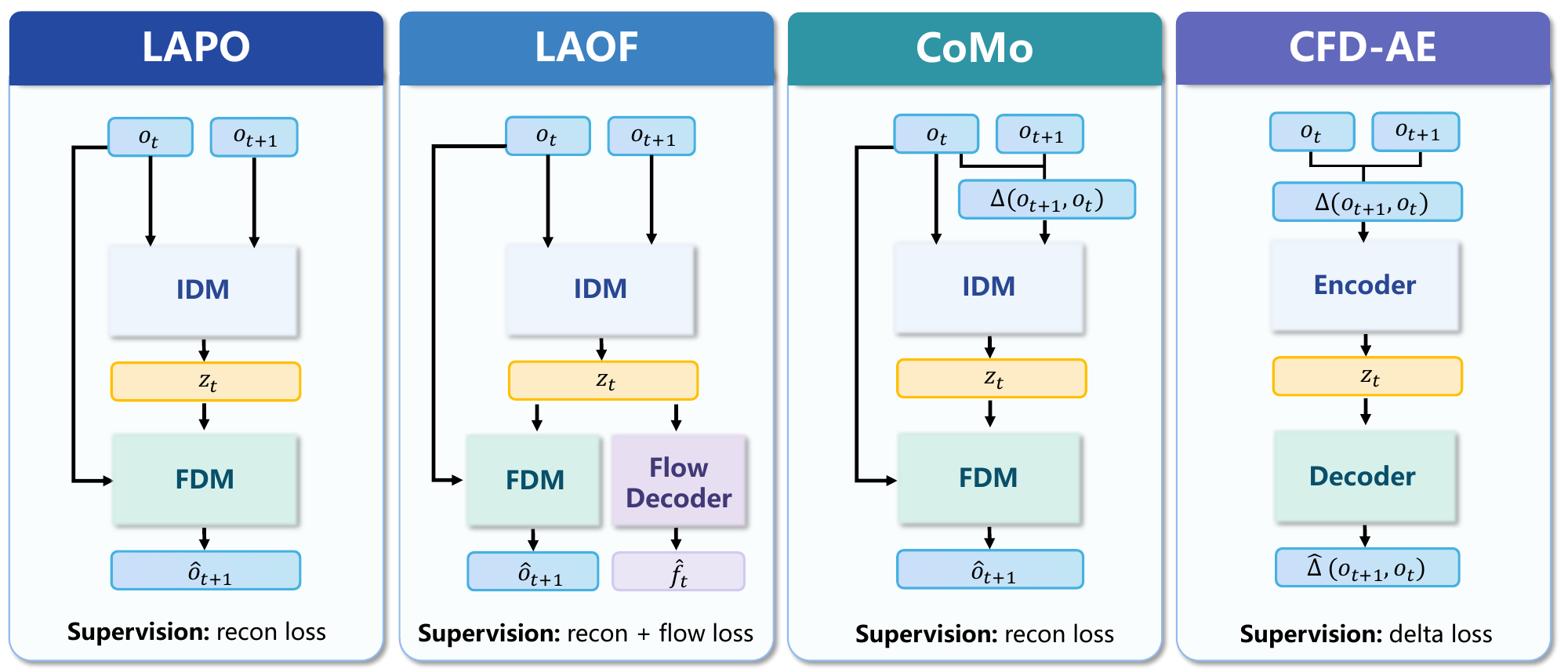}
    \caption{\textbf{Comparison of different latent action modeling paradigms.} LAPO, LAOF, and CoMo follow IDM-FDM to predict future frames, while CFD-AE reconstructs consecutive-frame differences. Arrows indicate information flow, with each panel's training losses shown at the bottom. Here, $z_t$ denotes the latent action and $\Delta$ consecutive-frame differences; hats indicate predictions.
    }
    \label{fig:cfm}
\end{figure}

$\Delta$RGB and $\Delta$DINO are computed as frame differences in pixel space and in the semantic feature space of DINOv2 \citep{oquab2023dinov2}, respectively, while optical flow is estimated using RAFT \citep{teed2020raft} or SEA-RAFT \citep{wang2024sea}. This setup allows us to assess whether semantic feature differences are more informative than raw pixel differences, and further examine the impact of optical flow quality. All methods in Fig.~\ref{fig:cfm} share the same autoencoding pipeline, ensuring that they are evaluated under the same latent action dimensionality for a fair comparison.

\textbf{a) LAPO} \citep{schmidt2024learning}: It is the standard self-supervised IDM-FDM framework for latent action learning. Given consecutive frames $(o_t,o_{t+1})$, the IDM infers a latent action $z_t$, and the FDM predicts the future frame conditioned on the current frame and the inferred latent action:
\begin{equation}
    z_t=\mathrm{IDM}(o_t, o_{t+1}),
    \quad
    \hat{o}_{t+1}=\mathrm{FDM}(o_t, z_t).
\end{equation}
The two models are jointly optimized in a self-supervised manner using the future-frame reconstruction objective
\begin{equation}
    \mathcal{L}_{\mathrm{LAPO}}
    =
    \mathcal{L}_{\mathrm{recon}},
    \quad
    \mathcal{L}_{\mathrm{recon}}
    =
    \| o_{t+1} - \hat{o}_{t+1} \|_2^2 .
\end{equation}
By minimizing this objective, the inferred latent action is encouraged to capture the transition information necessary to explain how the current frame evolves into the future frame, without requiring ground-truth physical actions.

\textbf{b) LAOF} \citep{bu2026laof}: It extends the standard IDM-FDM framework by introducing optical flow as an auxiliary supervision signal for latent action learning. The motivation is that optical flow explicitly represents pixel-wise motion between consecutive frames, encouraging latent actions to focus on action-related dynamics rather than environmental distractors. Besides predicting the future frame with the FDM, LAOF also introduces a FlowDecoder to predict the target optical flow from the latent action:
\begin{equation}
    \hat{f}_{t}=\mathrm{FlowDecoder}(z_t).
\end{equation}
The overall training objective combines future-frame reconstruction with optical-flow reconstruction
\begin{equation}
    \mathcal{L}_{\mathrm{LAOF}}
    =
    \mathcal{L}_{\mathrm{recon}}
    +
    \mathcal{L}_{\mathrm{flow}},
    \quad
    \mathcal{L}_{\mathrm{flow}}
    =
    \| f_t - \hat{f}_t \|_2^2,
\end{equation}
where $f_t$ denotes the optical flow estimated from consecutive frames $(o_t,o_{t+1})$ by an off-the-shelf optical flow estimator.

\textbf{c) CoMo} \citep{yang2026learning}: It is designed to mitigate shortcut learning caused by the causal leakage issue in the standard IDM-FDM framework by replacing the future frame input with the consecutive-frame differences. Specifically, the IDM infers the latent action from the current frame and the consecutive-frame differences $\Delta_t$, while the FDM reconstructs the future frame from the current frame and the inferred latent action:
\begin{equation}
    z_t=\mathrm{IDM}(o_t, \Delta_t),
    \quad
    \hat{o}_{t+1}=\mathrm{FDM}(o_t, z_t).
\end{equation}
Following the original CoMo setting~\citep{yang2026learning}, we instantiate $\Delta_t$ as $\Delta$DINO. The training objective remains the standard future-frame reconstruction loss $\mathcal{L}_{\mathrm{CoMo}}=\mathcal{L}_{\mathrm{recon}}$.

\textbf{d) CFD-AE}: In contrast to the IDM-FDM framework, CFD-AE directly learns latent actions by compressing and reconstructing the consecutive-frame differences $\Delta_t$, without modeling future-frame generation:
\begin{equation}
    z_t=\mathrm{Encoder}(\Delta_t),
    \quad
    \hat{\Delta}_t=\mathrm{Decoder}(z_t).
\end{equation}
The training objective is simply to reconstruct the input differences representation:
\begin{equation}
    \mathcal{L}_{\mathrm{CFD-AE}}
    =
    \mathcal{L}_{\mathrm{delta}},
    \quad
    \mathcal{L}_{\mathrm{delta}}
    =
    \| \Delta_t - \hat{\Delta}_t \|_2^2 .
\end{equation}
\subsection{Learning Objectives and Regularization}
The overall optimization objective is formulated as
\begin{equation}
\mathcal{L}=\mathcal{L}_{\mathrm{method}}+\lambda_{\mathrm{reg}}\mathcal{L}_{\mathrm{reg}}
\end{equation}
where $\mathcal{L}_{\mathrm{method}}$ denotes the learning objective defined by the latent action modeling method (e.g., LAPO, LAOF, CoMo, or CFD-AE), $\mathcal{L}_{\mathrm{reg}}$ denotes the latent action representation regularization term, and $\lambda_{\mathrm{reg}}$ is the regularization strength that controls the trade-off between reconstruction fidelity and latent representation regularity.

\textbf{a) AE} \citep{bank2023autoencoders}: It serves as the simplest baseline, where the inverse dynamics model deterministically encodes the input into a latent action $z_t$ without any explicit latent action regularization. The model is trained solely with the method-specific objective $\mathcal{L}_{\mathrm{method}}$.

\textbf{b) VQ-VAE} \citep{van2017neural}: It introduces a discrete latent action regularization on the latent action inferred by the inverse dynamics model. Specifically, the latent action $z_t$ is quantized by assigning it to its nearest codebook embedding $e_k$. The corresponding vector-quantization regularization is defined as
\begin{equation}
\mathcal{L}_{\mathrm{reg}}^{\mathrm{VQ}}
=
\| \mathrm{sg}[z_t]-e_k \|_2^2
+
\beta_{\mathrm{vq}}
\| z_t-\mathrm{sg}[e_k] \|_2^2,
\end{equation}
where $\mathrm{sg}[\cdot]$ denotes the stop-gradient operator and $\beta_{\mathrm{vq}}$ is the commitment weight. By constraining latent actions to a finite set of learnable codebook embeddings, VQ-VAE encourages compact, discrete, and reusable latent representations.

\textbf{c) VAE} \citep{kingma2013auto}: It introduces a KL divergence regularization to constrain the conditional latent distribution $p_{\mathrm{IDM}}(z_t \mid o_t, o_{t+1})$ toward a predefined prior distribution. The corresponding regularization objective is
\begin{equation}
    \mathcal{L}_{\mathrm{reg}}^{\mathrm{VAE}}
    =
    D_{\mathrm{KL}}
    \big(
    p_{\mathrm{IDM}}(z_t \mid o_t,o_{t+1})
    \,\|\,
    q(z_t)
    \big),
\end{equation}
where $q(z_t)=\mathcal{N}(0,I)$ denotes a standard Gaussian prior. By minimizing the KL divergence between the inferred latent distribution and the prior, VAE encourages a smooth, continuous, and well-structured latent space.

\textbf{d) Sparsity} \citep{garrido2026learning}: Its regularization encourages each latent action to activate only a small subset of latent dimensions, resulting in more compact and interpretable latent representations. Following \citep{garrido2026learning}, we combine sample-level sparsity regularization with Variance-Covariance-Mean (VCM) regularization to avoid degenerate solutions. Given a minibatch of latent actions $Z\in\mathbb{R}^{N\times d_z}$, where each row corresponds to a latent action $z_t$, the regularization objective is defined as
\begin{equation}
    \mathcal{L}_{\mathrm{reg}}^{\mathrm{Sparse}}
    =
    \mathcal{L}_{\mathrm{VCM}}
    +
    \frac{1}{N}
    \sum_{i=1}^{N}
    E(Z_i),
\end{equation}
where
\begin{equation}
    E(z_t)
    =
    \lambda_{\ell_2}
    \max\!\left(
    \sqrt{d_z}-\|z_t\|_2^2,
    0
    \right)
    +
    \lambda_{\ell_1}\|z_t\|_1.
\end{equation}
The VCM regularization is given by
\begin{align}
    \mathcal{L}_{\mathrm{VCM}}
    =
    &\lambda_V
    \frac{1}{d_z}
    \sum_{d=1}^{d_z}
    \max\!\left(
    1-\sqrt{\operatorname{Var}(Z_{.,d})},
    0
    \right)
    \nonumber\\
    &+
    \lambda_C
    \frac{1}{d_z(d_z-1)}
    \sum_{i\neq j}
    \operatorname{Cov}(Z)_{i,j}^{2}
    \nonumber\\
    &+
    \lambda_M
    \frac{1}{Nd_z}
    \sum_{i=1}^{N}
    \sum_{j=1}^{d_z}
    Z_{i,j}.
\end{align}
This regularization encourages sparse latent action representations while preserving sufficient variance, reducing redundancy across latent dimensions, and preventing representation collapse. In practice, we follow the regularization hyperparameters in \citep{garrido2026learning} and use $\lambda_{\ell_1}=0.01, \lambda_{\ell_2}=1$, $\lambda_V=0.1$, $\lambda_C=0.001$, and $\lambda_M=0.1$.

\textbf{e) SIGReg} \citep{balestriero2025lejepa}: It introduces statistical regularization of latent action representations by encouraging the latent action distribution to match an isotropic Gaussian target distribution. Directly assessing normality in high-dimensional spaces is statistically challenging, as most classical normality tests are designed for univariate distributions. Following \citep{balestriero2025lejepa}, SIGReg addresses this issue by projecting latent actions onto multiple random one-dimensional directions and matching the resulting projected distributions to a standard Gaussian. Given a minibatch of latent actions $Z$, SIGReg samples $M$ random unit-norm directions $u^{(m)}\in\mathbb{S}^{d_z-1}$ and computes the projected representations $h^{(m)}=Zu^{(m)},\ m=1,\ldots,M.$ The corresponding regularization term is defined as
\begin{equation}
    \mathcal{L}_{\mathrm{reg}}^{\mathrm{SIGReg}}
    =
    \frac{1}{M}
    \sum_{m=1}^{M}
    T\!\left(h^{(m)}\right),
\end{equation}
where $T(\cdot)$ denotes the univariate Epps--Pulley statistic \citep{epps1983test}, which measures the discrepancy between each projected distribution and a standard Gaussian. Motivated by the Cramér--Wold theorem \citep{cramer1936some}, SIGReg encourages an isotropic Gaussian latent distribution by matching its projections along sampled random directions to a standard Gaussian. Consequently, SIGReg encourages diverse, collapse-resistant latent action representations while regularizing their distribution toward an isotropic Gaussian.

\subsection{Latent Action Integration Strategies}
Given the VLM backbone $\mathrm{VLM}_{\psi}$ fine-tuned with latent actions, the current observation $o_t$ and language instruction $l$ are first encoded into a high-level representation $h_t=\mathrm{VLM}_{\psi}(o_t,l)$, which is then passed to an action head to predict the physical action $\hat{a}_t=\mathrm{ActionHead}(h_t)$, where $\psi$ denotes the parameters of the VLM backbone. The corresponding latent action prediction loss $\mathcal{L}_{\mathrm{latent}}$ and physical action prediction loss $\mathcal{L}_{\mathrm{act}}$ are defined as
\begin{equation}
\mathcal{L}_{\mathrm{latent}}
=
\|z_t-\hat{z}_t\|_2^2,
\quad
\mathcal{L}_{\mathrm{act}}
=
\|a_t-\hat{a}_t\|_2^2.
\end{equation}

The three key design dimensions investigated in this work are summarized in Table~\ref{tab:design_choices}.
\input{sec/design_choices}

\textbf{a) Direct Action Prediction (DAP), $\mathrm{VLM} \rightarrow a_t$.}
During Stage II, a latent action head is attached to the VLM backbone to predict latent actions from the high-level feature representation, where $\hat{z}_t=\mathrm{ActionHead}_{z}(h_t)$. The VLM backbone together with the latent action head is optimized using $\mathcal{L}_{\mathrm{latent}}$, yielding a fine-tuned VLM backbone denoted by $\mathrm{VLM}_{\psi^*}$. During Stage III, the latent action head is removed and replaced with a physical action head. Given $h_t=\mathrm{VLM}_{\psi^*}(o_t,l)$, the physical action is predicted as
\begin{equation}
\hat{a}_t=\mathrm{ActionHead}_{a}(h_t).
\end{equation}
The model is then optimized using $\mathcal{L}_{\mathrm{act}}$ on robot action data.

\textbf{b) Latent-to-Action Prediction (LAP), $\mathrm{VLM} \rightarrow z_t \rightarrow a_t$.}
Stage II is the same as in DAP. During Stage III, the fine-tuned VLM backbone and the latent action head are inherited from Stage II. Given $h_t=\mathrm{VLM}_{\psi^*}(o_t,l)$, instead of directly predicting the physical action from $h_t$, the model first predicts a latent action and subsequently decodes it into the physical action:
\begin{equation}
\hat{z}_t=\mathrm{ActionHead}_{z}(h_t),
\quad
\hat{a}_t=\mathrm{ActionHead}_{a}(\hat{z}_t).
\end{equation}
The VLM backbone together with the latent action head and physical action head is jointly optimized using $\mathcal{L}_{\mathrm{act}}$ on robot action data.

\textbf{c) Joint Latent-Action Prediction (JAP), $\mathrm{VLM} \rightarrow [z_t,a_t]$.}
A shared action head simultaneously predicts latent and physical actions:
\begin{equation}
[\hat{z}_t,\hat{a}_t]
=
\mathrm{ActionHead}_{z,a}(h_t).
\end{equation}
Unlike DAP and LAP, which first fine-tune the VLM backbone using latent actions before downstream policy learning, JAP assumes access to robot action data and jointly optimizes latent and physical action prediction in a single training stage. Therefore, no separate latent-action-only VLM fine-tuning stage is required. During Stage III, the VLM backbone together with the joint action head is optimized using the objective $\mathcal{L}=\mathcal{L}_{\mathrm{latent}}+\mathcal{L}_{\mathrm{act}}$, where $\mathcal{L}_{\mathrm{latent}}$ is computed on video data and $\mathcal{L}_{\mathrm{act}}$ is computed on robot action data.

\textbf{d) JAP-DAP and JAP-LAP.} The VLM backbone is first jointly fine-tuned with both latent actions and physical actions following the JAP setting. The trained action head is then discarded, and a new DAP or LAP head is trained from scratch. Both the joint fine-tuning and subsequent training with the new action head take place in Stage III.

%% file: sec/design_choices.tex
\begin{table}[t!]
\centering
\caption{Design choices explored in this work. Columns list latent action modeling paradigms, learning objectives and regularization, and latent action integration strategies. Design III distinguishes whether physical action annotations are accessible during VLM backbone fine-tuning.}
\label{tab:design_choices}

\scriptsize
\renewcommand{\arraystretch}{1.6}

\setlength{\tabcolsep}{6pt}

\resizebox{0.95\linewidth}{!}{
\begin{tabular}{@{\hspace{6pt}}l@{\hspace{10pt}}l@{\hspace{10pt}}l@{\hspace{6pt}}}
\hline

\begin{tabular}[c]{@{}c@{}}
\textbf{Design I} \\
(\circB: IDM-FDM \ \circC: CFD-AE)
\end{tabular}
&
\begin{tabular}[c]{@{}c@{}}
\textbf{Design II} \\
(\circB: Discrete \ \circC: Continuous)
\end{tabular}
&
\begin{tabular}[c]{@{}c@{}}
\textbf{Design III} \\
(\circB: Inaccessible \ \circC: Accessible)
\end{tabular}
\\

\hline

\hspace{5pt}\circB\ LAPO\citep{schmidt2024learning}     & \hspace{20pt} \circC\ AE\citep{bank2023autoencoders}  &  \hspace{30pt}\circB\ DAP      \\
\hspace{5pt}\circB\ LAOF\citep{bu2026laof}              & \hspace{20pt} \circB\ VQ-VAE\citep{van2017neural}      & \hspace{30pt}\circB\ LAP      \\
\hspace{5pt}\circB\ CoMo\citep{yang2026learning}        & \hspace{20pt} \circC\ VAE\citep{kingma2013auto}        & \hspace{30pt}\circC\ JAP      \\
\hspace{5pt}\circC\ RAFT\citep{teed2020raft} / SEA-RAFT\citep{wang2024sea} & \hspace{20pt}  \circC\ Sparsity\citep{garrido2026learning}  & \hspace{30pt}\circC\ JAP-DAP  \\
\hspace{5pt}\circC\ $\Delta$RGB / $\Delta$DINO\citep{oquab2023dinov2}     & \hspace{20pt} \circC\ SIGReg\citep{balestriero2025lejepa}   &  \hspace{30pt}\circC\ JAP-LAP  \\

\hline
\end{tabular}
}

\end{table}

%% file: sec/experiments_appendix.tex
\section{Experiments}
\label{app:experiments}
This appendix provides experimental details and supplementary discussion for Section~\ref{sec:experiments}.

\subsection{Experimental Setup}
\label{app:experimental_setup}
\noindent\textbf{Datasets.} The corpus described in Section~\ref{sec:experimental_setting} consists of a subset of the open-source Open X-Embodiment (\texttt{OXE})\citep{o2024open} real-world robot datasets, including \texttt{DROID} \citep{khazatsky2024droid} (45.6\%), \texttt{Language Table} \citep{lynch2023interactive} (11.7\%), \texttt{BC-Z} \citep{jang2021bc} (9.4\%), \texttt{Furniture Bench} \citep{heo2023furniturebench} (6.7\%), \texttt{Fractal} \citep{brohan2022rt} (6.7\%), \texttt{Bridge} \citep{walke2023bridgedata} (3.3\%), \texttt{FMB} \citep{luo2024fmb} (1.7\%), and \texttt{Stanford Hydra} \citep{belkhale2023hydra} (0.6\%), together with the simulated datasets \texttt{Robotwin} (10.6\%) and \texttt{liberoplus} (3.9\%). The \texttt{liberoplus} videos used in Stages I and II cover all perturbation types, including \texttt{RobotInit}. For the RoboTwin2.0 benchmark, we use the complete \texttt{Robotwin} dataset, which contains 50 bimanual manipulation tasks collected on the Aloha-AgileX robot platform, with each task including 50 clean demonstrations and 500 randomized demonstrations. For the LIBERO and LIBERO-Plus benchmarks, Stage-III policy training uses only the \texttt{Libero} datasets, including \texttt{Libero-10}, \texttt{Libero-goal}, \texttt{Libero-object}, and \texttt{Libero-spatial}, and the policies are evaluated on LIBERO-Plus under \texttt{RobotInit} without further Stage-III training. Notably, to ensure a fair comparison using the same training data, we compute optical flow directly from the raw data. In particular, we retain optical flow estimates for the \texttt{Noise} setting in LIBERO-Plus without applying any \mbox{setting-specific filtering.}

\textbf{LIBERO}\citep{liu2023libero}. It is a single-arm manipulation benchmark designed to evaluate compositional generalization and long-horizon task execution over shared objects and environments. We evaluate on 4 task suites, namely \texttt{\{Goal, Object, Spatial, and Long\}}, which respectively test generalization over task goals, manipulated objects, spatial relations, and temporally extended \mbox{instruction following.}

\textbf{LIBERO-Plus}\citep{fei2025libero}. It extends LIBERO by introducing diverse robustness perturbations to evaluate whether learned policies overfit to specific environmental conditions. It includes seven perturbation types: \texttt{\{Camera, RobotInit, Noise, Language, Layout, Light, Background\}}. The evaluation protocol in Section~\ref{sec:experimental_setting} evaluates policy robustness to variations in robot initialization while preserving the original \mbox{task objectives and evaluation setup.}

\textbf{RoboTwin2.0}\citep{chen2025robotwin}. It is a dual-arm manipulation benchmark designed to evaluate cross-domain generalization in bimanual settings. The benchmark provides paired environments with similar task structures but different visual appearances and dynamics. Due to computational constraints, we evaluate on 12 representative tasks with a horizon of 400 steps, selected from the full set of 50 tasks. These tasks include 6 easier tasks and 6 more challenging tasks for controlled ablations. Specifically, the selected tasks are \texttt{\{Adjust Bottle, Lift Pot, Move Playingcard Away, Move Stapler Pad, Pick Diverse Bottles, Place Container Plate, Place Mouse Pad, Place Object Stand, Place Phone Stand, Press Stapler, Rotate QRCode, and Turn Switch\}}.

\noindent\textbf{Implementation Details.} The IDM is composed of 24 encoder blocks for latent action inference, while the FDM is composed of 24 decoder blocks for future-frame prediction. To capture consecutive-frame motion representations across multiple temporal scales, input videos are temporally downsampled using a random factor sampled from \(\{1,2,3,4,5\}\). All video frames are center-cropped and resized to a fixed resolution of $224\times224$. We optimize the model using AdamW \citep{loshchilov2017decoupled}, with detailed training hyperparameters provided in Table~\ref{tab:training_hyperparameters}.

\begin{table}[t!]
\centering
\caption{Training hyperparameters for the three training stages. Stage I trains LAM, Stage II fine-tunes the VLM backbone with latent actions, and Stage III trains the downstream policy. Superscripts identify benchmark-specific settings.}
\label{tab:training_hyperparameters}
\begin{minipage}{0.72\linewidth}
\centering
\scriptsize
\setlength{\tabcolsep}{2pt}
\renewcommand{\arraystretch}{1.25}
\resizebox{\linewidth}{!}{%
\begin{tabular}{@{}lccc@{}}
\toprule
Hyperparameter & Stage I & Stage II & Stage III \\
\midrule
Base LR & $2.5\times10^{-5}$ & $2.5\times10^{-5}$ & $2.5\times10^{-5}$ \\
Action Head LR & -- & $1\times10^{-4}$ & $1\times10^{-4}$ \\
Qwen3-VL-4B LR & -- & $1\times10^{-5}$ & $1\times10^{-5}$ \\
LR Scheduler & Constant & Cosine with Min LR & Cosine with Min LR \\
Weight Decay & $1\times10^{-2}$ & $1\times10^{-8}$ & $1\times10^{-8}$ \\
Global Batch Size & 256 & 256 & $128^{+}$ / $32^{-}$ / $32^{*}$ \\
Training Steps & 150k & 150k & $80\text{k}^{+}$ / $150\text{k}^{-}$ / $40\text{k}^{*}$ \\
Action Chunk Size & - & 1 & $8^{+}$ / $50^{-}$ / $50^{*}$ \\
\bottomrule
\end{tabular}
}
\end{minipage}
\par\smallskip
{\small \mbox{LR: Learning Rate; $^{+}$: LIBERO / LIBERO-Plus; $^{-}$: RoboTwin2.0; $^{*}$: Real-World.}\par}
\end{table}

\subsection{\questionone~(\textbf{Q1})}
\label{app:q2_cfm}
We provide further discussion of the comparisons in Section~\ref{sec:q2_cfm}. Although $\Delta$RGB and optical flow both capture changes between consecutive frames, $\Delta$RGB directly encodes the pixel-wise difference between the two frames, whereas optical flow further transforms these changes into pixel-wise displacements, imposing a stronger motion-centric inductive bias. Such a representation may discard visual cues relevant to transition dynamics, including contact, occlusion and disocclusion, local deformation, and object boundary changes, which cannot always be adequately characterized by pixel-wise displacements. Moreover, optical flow introduces an additional motion estimation process, through which estimation errors may propagate to the learned latent action representation and reduce its robustness to visual perturbations. Visual perturbations such as noise may also affect optical flow estimation. Although SEA-RAFT achieves better average performance than RAFT on LIBERO, RAFT outperforms SEA-RAFT on LIBERO-Plus under distribution shifts. This comparison suggests that the choice of optical flow estimator can affect downstream performance \mbox{differently across benchmarks.}

Although $\Delta$RGB naturally suppresses static backgrounds by differencing consecutive frames in pixel-wise space, it contains little semantic information about the underlying visual state transitions. In contrast, $\Delta$DINO computes frame differences in a pretrained semantic feature space, where low-level visual details are partially abstracted into higher-level representations of objects and scenes, making it more suitable for robotic manipulation.

\subsection{\questiontwo~(\textbf{Q2})}
\label{app:q3_regularization}

\paragraph{Fig.~\ref{fig:lambda_selection}.}
We select the regularization strength $\lambda_{reg}$ using a composite metric defined as $\mathrm{SSIM\ Gain} + \mathrm{MSE\ Gain} - \mathrm{MLP} - \mathrm{Linear}$, with the chosen value indicated by the red vertical lines. Excessively strong regularization suppresses transition dynamics, leading to degraded future-frame reconstruction and weaker physical action prediction. This trend is particularly evident in the FDM reconstruction metrics, where overly large regularization strengths consistently reduce both $\mathrm{SSIM\ Gain}$ and $\mathrm{MSE\ Gain}$ for VAE, Sparsity, and SIGReg. Meanwhile, the probing metrics, $-\mathrm{Linear\ probe}$ and $-\mathrm{MLP\ probe}$, also deteriorate under inappropriate regularization strengths.

For the LIBERO-Plus result in Section~\ref{sec:q3_regularization}, a possible explanation is that the discrete bottleneck encourages the model to capture reusable high-level action primitives rather than overfit to fine-grained continuous variations in the source domain. Such action primitives may transfer more effectively under environmental perturbations, which is particularly beneficial for zero-shot evaluation. This advantage does not generalize across benchmarks. VQ-VAE achieves comparable performance on LIBERO but lower average performance on RoboTwin2.0 than the other evaluated methods. While discretization may improve robustness to distribution shifts, it also reduces the continuity and flexibility of latent actions, making it more difficult to further improve performance on benchmarks where performance is already saturated. Sparsity introduces additional hyperparameters and \mbox{SIGReg} incurs additional computational overhead from sampling random projection directions. These considerations support the default choice of VAE in Section~\ref{sec:q3_regularization}.

\subsection{\questionthree~(\textbf{Q3})}
\label{app:q4_action_head}

We supplement the overall comparison in Section~\ref{sec:q4_action_head} with the following results from Table~\ref{tab:main_results}.

\noindent\textbf{Physical Action Inaccessible.} The two exceptions to LAP outperforming DAP are LAOF (0.6563 vs. 0.6873) and VQ-VAE (0.7477 vs. 0.7560), with each value averaged over the three benchmarks. These exceptions show that retaining latent actions as an intermediate representation does not improve physical action prediction for every latent action configuration.

The comparison with JAP in Section~\ref{sec:q4_action_head} concerns overall mean performance across the evaluated configurations. The similar overall means do not imply that LAP and JAP perform similarly for each configuration. Averaging over the three benchmarks, LAP performs better than JAP with $\Delta$DINO and VAE, while JAP performs better with RAFT and LAOF. This distinction is relevant when choosing a strategy for a specific latent action model.

\noindent\textbf{Physical Action Accessible.} Both JAP-DAP and JAP-LAP achieve higher mean success rates than JAP on each of the three benchmarks when averaged over the 12 configurations. Their relative performance differs between benchmarks. Among these three strategies, JAP-LAP achieves the highest mean success rate on LIBERO (0.8964) and LIBERO-Plus (0.4243), while JAP-DAP achieves the highest mean on RoboTwin2.0 (0.8450), compared with 0.8428 for JAP-LAP. These benchmark-specific comparisons supplement the overall ranking in Section~\ref{sec:q4_action_head} and explain why the preferred variant can depend on the target benchmark.

\begin{figure}[t!]
	\centering
    \includegraphics[width=0.95\linewidth]{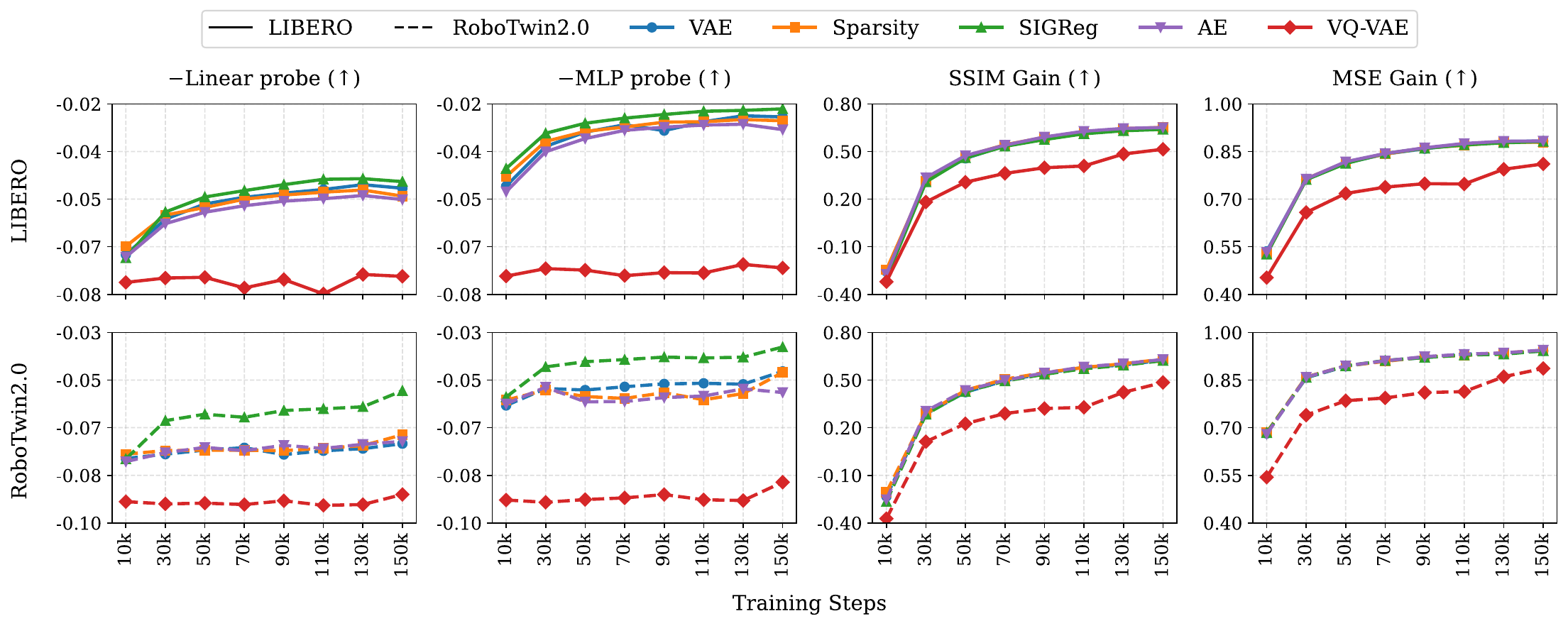}
    \caption{\textbf{Performance of different regularization methods throughout training.} The four columns plot the proxy metrics defined in Section~\ref{sec:q1_proxy_metrics} against training steps, up to 150K. The top and bottom rows show LIBERO (solid lines) and RoboTwin2.0 (dashed lines), respectively. Colors distinguish AE and the four regularization methods. Higher values indicate better proxy scores.}
    \label{fig:training_process}
\end{figure}

\subsection{\questionfour~(\textbf{Q4})}
\label{app:q1_proxy_metrics}

\paragraph{Fig.~\ref{fig:latentaction_metrics}.}

The \textit{Regularization methods} setting in Fig.~\ref{fig:latentaction_metrics} comprises 21 experimental configurations, including one AE configuration; four regularization strengths, $\lambda_{\mathrm{reg}}\in\{10^{-2},10^{-4},10^{-6},10^{-8}\}$, evaluated for each of VAE, Sparsity, SIGReg, and VQ-VAE; four additional configurations corresponding to the regularization strengths selected in this work, namely VAE ($10^{-7}$), Sparsity ($10^{-5}$), SIGReg ($10^{-3}$), and VQ-VAE ($1$). The \textit{+ Latent action dimensionality} setting further augments these configurations with eight configurations with different latent action dimensionalities, $d_z\in\{8,16,32,64,128,256,512,1024\}$, for 29 experimental configurations in total. For each configuration, we report results under DAP and LAP, as these two settings train the VLM backbone solely using latent actions without physical action supervision in Stage II, thereby providing a direct measure of the correlation between proxy metrics of latent actions and downstream robotic manipulation performance.

These correlations summarize agreement across the evaluated configurations and do not directly measure how accurately a proxy metric selects the best-performing model. In particular, a positive correlation does not guarantee that the highest proxy score corresponds to the highest downstream success rate. This distinction supports the screening recommendation in Section~\ref{sec:q1_proxy_metrics}, with final model selection based on downstream task evaluation.

As dimensionality increases, latent actions can encode progressively richer information, which may reduce probe prediction errors even when such gains in representational capacity do not translate into improved downstream performance. At the same time, higher dimensional representations may facilitate shortcut learning for frame reconstruction, thereby weakening the discriminative \mbox{power of reconstruction metrics.}

The lower probe losses on LIBERO discussed in Section~\ref{sec:q1_proxy_metrics} and shown in Fig.~\ref{fig:training_process} may partly reflect differences in action parameterization. However, LIBERO and RoboTwin2.0 also differ in action dimensionality and task distribution, and the comparability of their probe losses depends on action scaling and loss normalization. Latent actions derived from consecutive-frame motion may be more aligned with differential control formulations. In contrast, position control requires predicting or recovering absolute target states, which may be more task-dependent and may contribute to higher probe losses. The four proxy metrics generally improve over training. This trend suggests that the learned latent action representations progressively encode more action-relevant information, even under joint position control, potentially indicating the emergence of a more structured motion representation space during training.

\subsection{\questionfive~(\textbf{Q5})}
\label{app:q5_latent_action_dimension}

\paragraph{Fig.~\ref{fig:latentdims}.}
Since LIBERO is relatively prone to performance saturation and potential overfitting, the marginal differences reported in Section~\ref{sec:q5_latent_action_dimension} are unlikely to reflect meaningful differences in generalization. The LIBERO-Plus results suggest that lower-dimensional latent spaces may help suppress redundant variations for relatively simple single-arm manipulation tasks. The RoboTwin2.0 results suggest that excessively small latent dimensions are insufficient to capture the complexity of bimanual manipulation, whereas excessively large latent spaces do not consistently improve performance and may instead introduce redundant or less structured representations.

\begin{figure}[t!]
	\centering
    \includegraphics[width=0.5\linewidth]{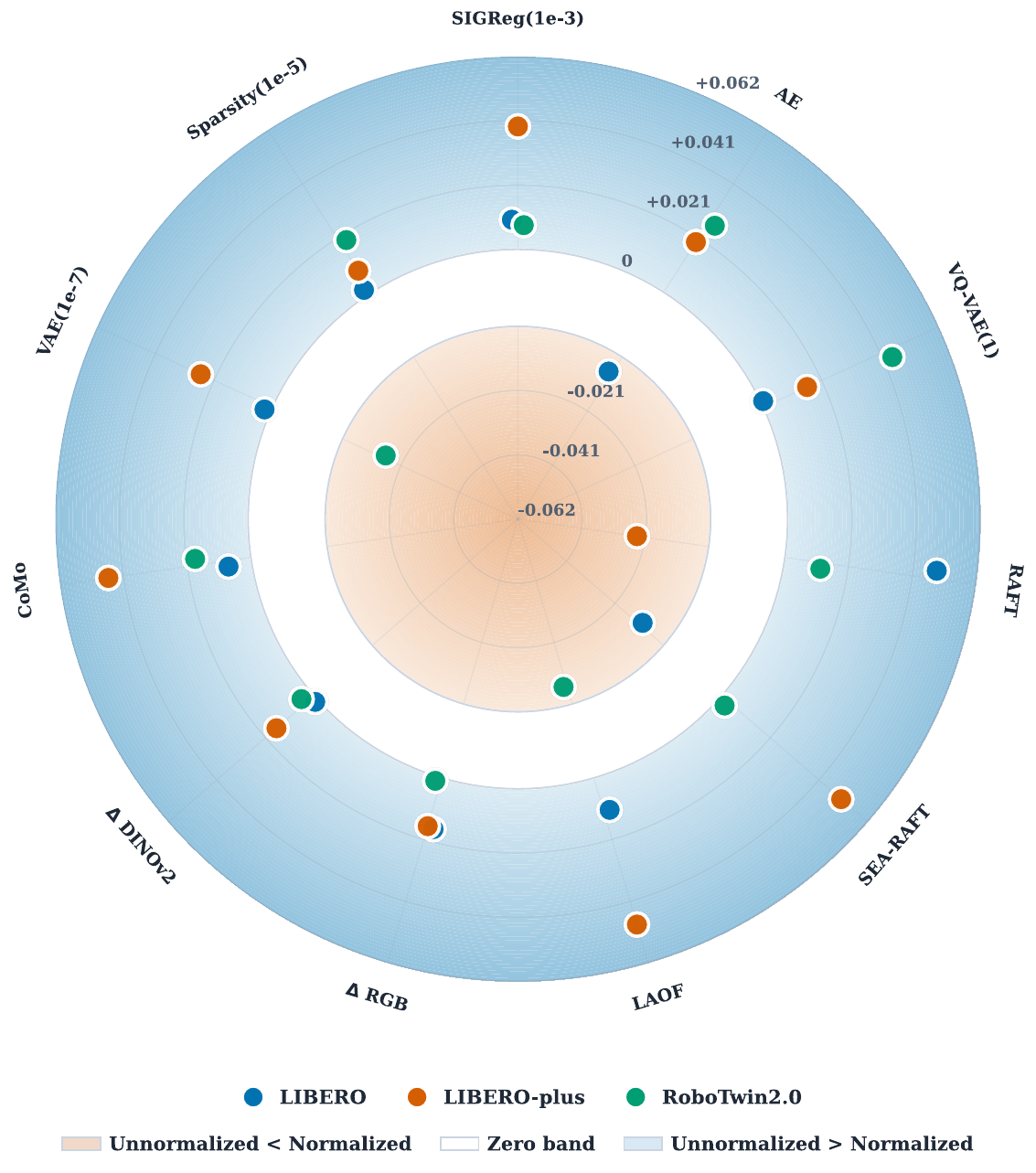}
    \caption{\textbf{Performance difference between unnormalized and normalized settings across methods and benchmarks.} Angular groups identify methods, with regularization strengths in parentheses; radial position shows the success-rate gap $(\mathrm{Unnormalized} - \mathrm{Normalized})$, averaged over DAP and LAP. Blue, orange, and green points denote LIBERO, LIBERO-Plus, and RoboTwin2.0, respectively. The white annulus denotes the zero band; the outer blue and inner orange regions indicate gains and losses without normalization, respectively.}
    \label{fig:denorm}
\end{figure}

For the normalization comparison in Section~\ref{sec:q5_latent_action_dimension} and Fig.~\ref{fig:denorm}, a plausible explanation is that regularization has already imposed sufficient structure and stability on the latent space, which may limit the benefit of additional normalization; normalization may also suppress magnitude-related information relevant to downstream \mbox{policy learning.}

\subsection{\questionsix~(\textbf{Q6})}
\label{app:q6_scalelaw}

\paragraph{Fig.~\ref{fig:scalelaw}.}
With the Stage III training data, algorithm, model architecture, and optimization hyperparameters kept identical, this comparison examines how changing the Stage-II fine-tuning corpus affects the VLM backbone initialization for downstream policy learning. Specifically, the training data, algorithm, model architecture, and all optimization hyperparameters in Stage III are kept identical, while the Stage-II VLM backbone is fine-tuned using either \texttt{Robotwin}+\texttt{liberoplus} (14.5\% of the full dataset) or \texttt{Robotwin}+\texttt{liberoplus}+\texttt{OXE} (100\%). The improvements for both DAP and LAP reported in Section~\ref{sec:q6_scalelaw} show that the benefit is observed with both direct physical action prediction from VLM features and prediction through latent actions. Thus, the observed benefit is not restricted to retaining latent actions as an intermediate representation during Stage III.

\subsection{Real-World Experiments}
\label{app:real_world_experiments}

\paragraph{Fig.~\ref{fig:real_world_tasks}.}
Specifically, we design four representative tabletop manipulation tasks: \texttt{Close} \texttt{Drawer}, involving closing the top red drawer; \texttt{Open} \texttt{Drawer}, involving opening the top red drawer; \texttt{Stack} \texttt{Bowls}, requiring two bowls to be stacked; and \texttt{Place} \texttt{Block}, requiring a block to be placed into a bowl. We compare OpenVLA-OFT, which is trained using the original Qwen3-VL-4B backbone, and OpenVLA-OFT (LA-Tuned), which uses a Qwen3-VL-4B backbone fine-tuned with latent actions on 59M video frames annotated with latent actions using LAPO with VAE ($10^{-7}$) regularization, a latent action dimensionality of $d_z=32$, and unnormalized latent actions. For each task, 50 robot demonstrations are collected under joint position control. To ensure a fair comparison, both models follow the same Stage III training configuration (Table~\ref{tab:training_hyperparameters}) and are each trained on the same combined dataset of 200 robot demonstrations collected across the four manipulation tasks. Furthermore, OpenVLA-OFT (LA-Tuned) uses DAP as the latent action integration strategy, ensuring that the only difference between two models is the parameters of the VLM backbone. We evaluate checkpoints at 6k, 8k, 10k, 20k, and 40k training steps. Each checkpoint is evaluated over 20 independent trials, resulting in 100 evaluations per task for each method. All experiments use only a single front-view RGB camera as the visual input.

During evaluation, each policy is deployed on the real robot in a closed-loop manner, with task success rate serving as the primary evaluation metric. The rollouts start from randomized initial states, resetting the scene before every rollout. To further increase evaluation diversity, we randomize both the initial object configurations and the placement of task-irrelevant distractor objects on the tabletop, while ensuring that the same set of evaluation scenes is used for all methods. A rollout is considered successful only if the robot completes the target manipulation within \mbox{a predefined execution horizon.}

Specifically, take the \texttt{Open} \texttt{Drawer} task as an example, the robot arm must grasp the round handle with a relatively large UMI gripper. This requires the gripper to sufficiently wrap around the handle while avoiding collisions with the upper cabinet surface. OpenVLA-OFT (LA-Tuned) exhibits better performance in this task. During evaluation, we further observe that OpenVLA-OFT (LA-Tuned) exhibits stronger robustness to task-irrelevant distractors. For example, when objects from other tasks are placed in the scene, such as bowls appearing during the drawer-opening task, the LA-Tuned model can still focus on the task-relevant object and complete the intended manipulation. In contrast, the baseline OpenVLA-OFT is more likely to switch to an incorrect task mode and interact with the distractor object. This suggests that latent-action tuning not only improves manipulation success rates, but also enhances the task-conditioned grounding ability of the VLM backbone, enabling the policy to better distinguish task-relevant affordances from irrelevant \mbox{visual distractors.}